\documentclass{article}

\usepackage{microtype}
\usepackage{graphicx}
\usepackage{subcaption}
\usepackage{booktabs} 

\usepackage{hyperref}

\usepackage[preprint]{icml2026}

\usepackage{amsmath}
\usepackage{amssymb}
\usepackage{mathtools}
\usepackage{amsthm}

\usepackage[capitalize, noabbrev]{cleveref}
\crefformat{subsection}{\S#2#1#3}

\theoremstyle{plain}
\newtheorem{theorem}{Theorem}[section]
\newtheorem{proposition}[theorem]{Proposition}

\newtheorem{corollary}[theorem]{Corollary}
\theoremstyle{definition}

\theoremstyle{remark}

\usepackage[textsize=tiny]{todonotes}

\usepackage{lipsum}

\usepackage{xspace}

\usepackage[shortlabels]{enumitem}

\usepackage{multirow}

\usepackage{tcolorbox}
\usepackage{caption}
\usepackage{tikz}
\usetikzlibrary{arrows.meta}

\usepackage{algorithmic}

\usepackage{bbm}

\usepackage[table]{xcolor}
\newcommand{\hl}{\cellcolor{white!70!black!30}\bfseries}

\usepackage{listings}
\usepackage{minted}

\usepackage{makecell}

\newcommand{\Ell}{\mathcal{L}}
\newcommand{\EE}{\mathbb{E}}
\newcommand{\KL}{\mathbb{D}_{\mathrm{KL}}}
\newcommand{\argmax}{\operatornamewithlimits{\arg\max}}
\newcommand{\ngem}{\textnormal{n\textsc{gem}}\xspace}
\newcommand{\nll}{\textnormal{\textsc{nll}}\xspace}
\newcommand{\sgem}{\textnormal{s\textsc{gem}}\xspace}
\newcommand{\x}{\boldsymbol{x}}
\newcommand{\y}{\boldsymbol{y}}
\newcommand{\z}{\boldsymbol{z}}
\newcommand{\dif}{\mathrm{d}}

\newcommand{\sg}[1]{\left\lfloor#1\right\rfloor}

\let\oldSigma\Sigma
\renewcommand{\Sigma}{\boldsymbol\oldSigma}
\let\oldmu\mu
\renewcommand{\mu}{\boldsymbol\oldmu}
\let\oldpi\pi
\renewcommand{\pi}{\boldsymbol\oldpi}
\let\oldpsi\psi
\renewcommand{\psi}{\boldsymbol\oldpsi}
\let\oldsigma\sigma
\renewcommand{\sigma}{\boldsymbol\oldsigma}
\let\oldtheta\theta
\renewcommand{\theta}{\boldsymbol\oldtheta}

\begin{document}
    \twocolumn[
    \icmltitle{Learning Mixture Density via Natural Gradient Expectation Maximization}
    \icmltitlerunning{Learning Mixture Density via Natural Gradient Expectation Maximization}
    \icmlsetsymbol{equal}{*}
    \begin{icmlauthorlist}
        \icmlauthor{Yutao Chen}{1}
        \icmlauthor{Jasmine Bayrooti}{2}
        \icmlauthor{Steven Morad}{1}
    \end{icmlauthorlist}
    \icmlaffiliation{1}{Faculty of Science and Technology, University of Macau}
    \icmlaffiliation{2}{University of Cambridge}
    \icmlcorrespondingauthor{Steven Morad}{smorad@um.edu.mo}
    \icmlkeywords{Machine Learning, ICML} \vskip 0.3in ]

    \printAffiliationsAndNotice{} 

    \begin{abstract}
     Mixture density networks are neural networks that produce Gaussian mixtures to represent continuous multimodal conditional densities. Standard training procedures involve maximum likelihood estimation using the negative log-likelihood (\nll) objective, which suffers from slow convergence and mode collapse. In this work, we improve the optimization of mixture density networks by integrating their information geometry. Specifically, we interpret mixture density networks as deep latent-variable models and analyze them through an expectation maximization framework, which reveals surprising theoretical connections to natural gradient descent. We exploit such connections to derive the natural gradient expectation maximization (\ngem) objective. We empirically show that \ngem achieves up to 10$\times$ faster convergence while adding almost \emph{zero} computational overhead, and scales well to high-dimensional data where \nll otherwise fails.
\end{abstract}
    \section{Introduction}

Predictive modeling is a central task in probabilistic machine learning, where one posits a conditional distribution $p(\y|\x)$ over the  features $\x$ and targets $\y$. For cases where $\y$ is continuous, it is usually analytically convenient to assume that $p(\y|\x)$ is Gaussian. However, Gaussian distributions are unimodal and hence incapable of fitting \emph{multimodal} distributions. An alternative is the Gaussian mixture model, which approximates arbitrary continuous smooth densities via a finite convex combination of Gaussian components.

\begin{figure}[t]
    \centering
    \includegraphics[width=\linewidth]{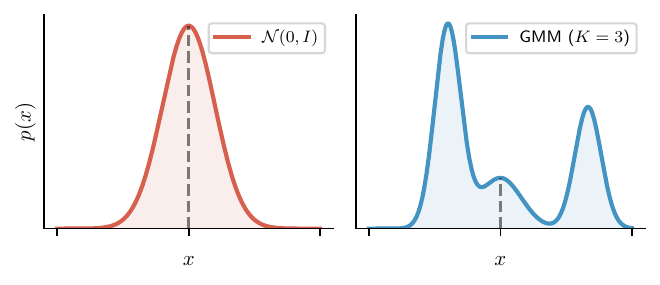}
    \caption{(\textbf{Left}) A standard Gaussian distribution compared with (\textbf{right}) a multimodal Gaussian mixture model. Both share the same mean $\EE[x]=0$ while the Gaussian mixture is more flexible. }
    \label{fig:1}
\end{figure}

Building upon Gaussian mixtures, mixture density networks \cite{bishop1994mixture} constitute a family of flexible probabilistic models designed for fitting complex and multimodal conditional distributions. Specifically, mixture density networks leverage neural networks to map features $\x$ to the distributional parameters of Gaussian mixtures that approximate the conditional density $p(\y|\x)$. Mixture density networks are conventionally optimized via stochastic gradient descent by minimizing the negative log-likelihood loss \cite{bishop1994mixture}.

Mixture density networks are widely used for uncertainty-aware prediction and multi-valued regression, where an input $\x$ may correspond to multiple valid outputs $\y$, e.g., speech synthesis \cite{capes2017siri}, inverse kinematics or forward dynamics \cite{ha2018world} in robotics. However, despite being accessible in principle, the optimization of mixture density networks is often challenging, impeding practical adoption. Particularly, the loss landscape of Gaussian mixture negative log-likelihood is non-convex and contains many poor local minima \cite{chen2024local}. Empirical studies \cite{li2019generating,makansi2019overcoming} also reveal that mixture density networks are susceptible to training inefficiency and mode collapse, where Gaussian mixtures effectively degenerate into one singular Gaussian.

\textbf{Our Contributions} \space  We revisit mixture density network optimization through the lens of latent-variable models and expectation maximization (EM). We identify theoretical connections between EM and natural gradient descent for Gaussian mixtures, and exploit such connections to construct a tractable natural gradient EM (\ngem) objective that yields fast and stable convergence by leveraging the information geometry of the underlying probabilistic models.

Empirically, we demonstrate that \ngem effectively stabilizes and accelerates mixture density network optimization, achieving up to 10$\times$ faster convergence with negligible extra computational overhead. We further validate the scalability of \ngem with high-dimensional and practical datasets.

    \section{Preliminaries}

In this section, we briefly review the foundational concepts: \hyperref[sec:2.1]{mixture density networks} (MDNs), \hyperref[sec:2.2]{natural gradient descent} (NGD), and \hyperref[sec:2.3]{expectation maximization} (EM), upon which the proposed method \ngem is established.

\phantomsection \label{sec:2.1}
\textbf{Mixture Density Networks} (MDNs) \cite{bishop1994mixture} model flexible conditional distributions $p(\y|\x)$ using Gaussian mixtures and neural networks. Specifically, MDNs learn a neural network $f(\x; \theta)$ parametrized by $\theta$ that maps conditions $\x$ to the parameters $\phi^{(\x)}$ of a Gaussian mixture, where
\begin{equation}
    \phi^{(\x)} = \{\pi_{k}^{(\x)}, \mu_{k}^{(\x)}, \Sigma_{k}^{(\x)}\}_{k=1}^{K} %
    \label{eq:1}
\end{equation}
is the collection of the weights\footnote{\space The Gaussian mixture weights $\pi$ satisfies $\pi\in\Delta^{K-1}$.}, means, and covariances of the $K$ Gaussian mixture components conditioned on $\x$. The conditional distribution $p(\y|\x)$ is thus
\begin{equation}
    p(\y|\x) = \sum_{k=1}^{K} %
    \pi_{k}^{(\x)}\mathcal{N}(\y;\mu_{k}^{(\x)}, \Sigma_{k}^{(\x)}). %
    \label{eq:2}
\end{equation}
Given data points $\{\x, \y\}$, the parameters $\theta$ are optimized typically by minimizing the negative log-likelihood (\nll)
\begin{equation}
    \Ell_{\nll}(\theta) = -\log \sum_{k=1}^{K} %
    \pi_{k}^{(\x)}\mathcal{N}(\y;\mu_{k}^{(\x)}, \Sigma_{k}^{(\x)}). %
    \label{eq:3}
\end{equation}

\phantomsection \label{sec:2.2}
\textbf{Natural Gradient Descent} (NGD) \cite{amari1998natural} is a generalization of the canonical gradient descent (GD) algorithm that accounts for the curvature of the parameter space of probabilistic models $p(\x|\theta)$. Consider minimizing the objective $J(\theta)$ with a learning rate $\beta$ via gradient descent
\begin{equation}
    \theta_{t+1} \leftarrow \theta_{t} - \beta \nabla J(\theta_{t}). %
    \label{eq:4}
\end{equation}
GD performs steepest descent with a Euclidean distance penalty between parameter updates $\Vert\theta_{t}-\theta_{t+1}\Vert_{2}^{2}$. For probabilistic models $p (\x|\theta)$, however, a more \emph{natural} penalty\footnote{\space We refer readers to \cref{app:a} for full justification for NGD.} is the KL divergence $\KL(p(\x|\theta_{t})\Vert p(\x|\theta_{t+1}))$ between the induced distributions, which leads to the NGD update:
\begin{equation}
    \theta_{t+1} \leftarrow \theta_{t} - \beta F^{-1}(\theta_{t})\nabla J(\theta_{t}), %
    \label{eq:5}
\end{equation}
where $F^{-1}(\theta_{t})$ is the inverse of the \emph{Fisher information matrix} (FIM) of $p(\x|\theta_{t})$. Specifically, the FIM $F(\theta)$, under certain conditions \citep[Section 5]{martens2020new}, can be defined as the negative expected Hessian of the log density
\begin{equation}
    F(\theta) = -\EE_{p(\x|\theta)}[\nabla^{2}_{\theta}\log p(\x|\theta)]. %
    \label{eq:6}
\end{equation}
Intuitively, NGD utilizes second-order derivatives (FIM) to capture the local geometry of probabilistic models, reprojecting the gradient updates appropriately for faster and more stable optimization. However, a critical downside of NGD lies in the computation and inversion of the FIM $F(\theta)$, which in many cases are intractable and require expensive approximation \citep{martens2020new}.

\phantomsection \label{sec:2.3}
\textbf{Expectation Maximization} (EM) \cite{dempster1977maximum} is an algorithm for finding maximum likelihood solutions of probabilistic models with missing data. EM is particularly well-known for its application in Gaussian mixture models, where each observed data point is assumed to be generated by one of the mixture components whose identity is missing.

Let $\x$ denote the observed data, $\z$ denote the missing data, and $p(\x ,\z|\theta)$ denote the probabilistic model parametrized by $\theta$. Directly maximizing the observed-data log-likelihood $\log p(\x|\theta) = \log \int p(\x,\z|\theta ) \dif \z$ is generally difficult due to the integration over $\z$. However, note that
\begin{equation}
    \log p(\x|\theta) %
    = \mathcal{F}(q(\z), \theta) %
    + \KL(q(\z) \Vert p(\z|\x;\theta )), %
    \label{eq:7}
\end{equation}
where
\begin{equation}
    \mathcal{F}(q(\z), \theta) %
    = \EE_{q(\z)}\left[\log \dfrac{p(\x,\z|\theta)}{q(\z)}\right] %
    \leq \log p(\x|\theta) %
    \label{eq:8}
\end{equation}
is a lower bound of $\log p(\x|\theta)$ (known as the \emph{evidence lower bound}, or ELBO) since KL divergence is non-negative \cite{neal1998view}.

The EM algorithm maximizes $\log p(\x|\theta)$ by instead maximizing the lower bound $\mathcal{F}(q(\z), \theta)$ iteratively. For each iteration $t$, we alternate between the E-step and the M-step:
\begin{itemize}
    \item (\textbf{E-step}) we maximize $\mathcal{F}(q(\z), \theta)$ with
        $\theta =\theta_{t}$
        \begin{equation}
            q_{t}(\z) %
            = \argmax_{q(\z)}\mathcal{F}(q(\z), \theta_{t}) %
            = p(\z|\x;\theta_{t}). %
            \label{eq:9}
        \end{equation}
    \item (\textbf{M-step}) we maximize $\mathcal{F}(q(\z), \theta)$ with
        $q(\z) =q_{t}(\z)$
        \begin{align}
            \kern-0.5em %
            \theta_{t+1} & = \argmax_{\theta}\mathcal{F}(q_{t}(\z), \theta) \notag \\
                         & = \argmax_{\theta}\int p(\z|\x;\theta_{t})\log p(\x,\z|\theta)\dif\z. %
            \label{eq:10}
        \end{align}
\end{itemize}
EM guarantees monotonic improvement of $\log p(\x|\theta)$ until convergence to some \emph{local} maxima \citep{dempster1977maximum}.

For cases where exact maximization is not tractable, the M-step can be substituted with gradient ascent steps, leading to the generalized EM \citep[Chapter~9]{bishop2006pattern} algorithm.

    \begin{figure*}[t]
    \centering
    \begin{tabular}{ccc}
    \multicolumn{2}{c}{%
        \includegraphics[width=0.5\textwidth]{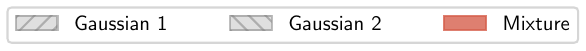}
    } & \\[-1.8em]
    \subfloat[\nll\label{fig:2a}]{%
        \includegraphics[width=0.27\textwidth]{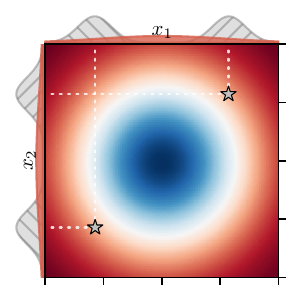}
    } &
    \subfloat[\ngem\label{fig:2b}]{%
        \includegraphics[width=0.27\textwidth]{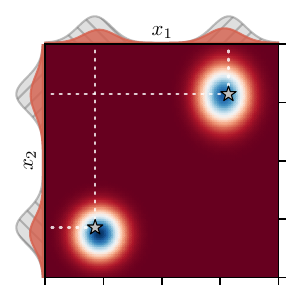}
    } &
    \subfloat[Convergence speed with $\beta=0.01$\label{fig:2c}]{%
        \includegraphics[width=0.37\textwidth]{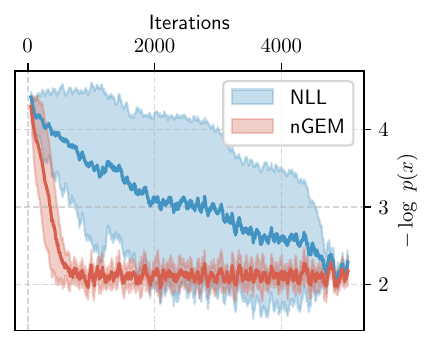}
    }
    \end{tabular}
    \caption{Fitting two Gaussians with a Gaussian mixture model (GMM) in $\mathbb{R}^2$. \emph{(a)} Mode collapse using \nll loss. \emph{(b)} Mode separation using \ngem loss. Heatmaps denote the probability density of the learned GMM, while stars ($\star$) denote means of the ground truth Gaussians. Marginals distributions are also displayed with hashed \textcolor{gray}{gray} regions representing the ground truth Gaussians and \textcolor{red}{red} regions representing the learned mixture density. \emph{(c)} Negative log-likelihoods ($\downarrow$) of GMMs with a learning rate $\beta=10^{-2}$, averaged ($\pm$ std) across 5 random seeds.}
    \label{fig:2}
\end{figure*}

\section{Natural Gradient EM}

While Gaussian mixtures in theory are capable of approximating any continuous density \citep[Chapter~2]{bishop2006pattern}, in practice mixture density networks are found to suffer from slow convergence and mode collapse \cite{li2019generating, makansi2019overcoming}, where the model manages to locate a \emph{subpar mean fit} using one mixture component to cover all or multiple modes in the target distribution (see \cref{fig:2a}).

One could consider improving the standard \nll objective using NGD, which accounts for the geometry of the parameter space. Unfortunately, this approach is not applicable directly as the FIM of Gaussian mixtures is non-analytical\footnote{\space We refer the readers to \cref{app:b5} for further justification.}.

To address this intractability, we propose natural gradient expectation maximization (\ngem). Building upon classic Gaussian mixture models, we establish an EM framework for learning mixture density (\cref{sec:3.1}). We show that EM, NGD, and \nll for Gaussian mixtures are connected and in fact equivalent under mild assumptions (\cref{sec:3.2}). We then exploit such connections to derive a \emph{tractable} natural gradient EM (\ngem) algorithm for learning mixture density (\cref{sec:3.3}).

\subsection{Stochastic Gradient EM}
\label{sec:3.1}

\newcommand{\joint}[1]{\pi_{#1}^{(\x)}\mathcal{N}(\y;\mu_{#1}^{(\x)},\Sigma_{#1}^{(\x)})}

We begin by reinterpreting mixture density networks as deep latent-variable models, with observed features $\x$ and targets $\y$, latent variables $\z$, and network parameters $\theta$. The joint conditional distribution of $\{ \y,\z\}$ given $\x$ is\footnote{\space We shall omit $\theta$ hereafter whenever there is no ambiguity.}
\begin{equation}
    p(\y,\z|\x;\theta) = p(\z|\x;\theta) p(\y|\x,\z;\theta),
    \label{eq:11}
\end{equation}
where $p(\z|\x)$ is a categorical distribution
\begin{equation}
    p(\z|\x) = \text{Cat}(\pi_{1}^{(\x)}, \pi_{2}^{(\x)}, \dots, \pi_{K}^{(\x)}),
    \label{eq:12}
\end{equation}
and $p(\y|\x,\z)$ is a Gaussian distribution
\begin{equation}
    p(\y|\x,\z=k)= \mathcal{N}(\y;\mu_{k}^{(\x)},\Sigma_{k}^{(\x)}).
    \label{eq:13}
\end{equation}
We obtain the observed-data log-likelihood of MDNs by marginalizing $p(\y, \z| \x )$ over latent variables $\z$:
\begin{align}
    \log p(\y|\x) = \log\sum_{k=1}^{K}\joint{k},
    \label{eq:14}
\end{align}
which coincides with the negative of $\Ell_{\nll}$ \eqref{eq:3}. Recall that EM maximizes the observed-data log-likelihood (or equivalently minimizing \nll) by maximizing the lower bound objective $\mathcal{F}(q(\z),\theta)$ \eqref{eq:8}. Similarly for MDNs, we follow an iterative EM process. In the \textbf{E-step}, we compute \eqref{eq:9}
\begin{align}
    q_{t}(\z=k|\x) & = p(\z=k|\x,\y;\theta_{t}) \notag \\
                   & = \dfrac{\joint{k}}{\sum_{j=1}^{K}\joint{j}}.
    \label{eq:15}
\end{align}
For brevity, we let $\rho_{k}^{(\x)}=q_{t}(\z=k|\x)$ where $\rho^{(\x)}\in\mathbb{R} ^{K}$ is known as the \emph{responsibilities}, explaining how much each mixture component is responsible for data points $\{\x,\y\}$.

In the \textbf{M-step}, however, we do not directly maximize w.r.t. the Gaussian mixture parameters $\phi^{(\x)}$ \eqref{eq:1}, but instead w.r.t. the parameters $\theta$ of the neural network $f(\x;\theta)$ that generates $\phi^{(\x)}$. Exact maximization is generally intractable due to the complex and non-linear nature of neural networks.

Following the generalized EM algorithm, we opt for stochastic gradient descent to maximize the M-step objective \eqref{eq:10}, leading to the stochastic gradient EM (\sgem) loss
\begin{align}
    \kern-0.5em %
    \Ell_{\sgem}(\theta)
    & = - \int p(\z|\x,\y;\theta_{t}) \log p(\y,\z|\x;\theta) \dif\z \notag \\
    & = - \sum_{k=1}^{K}\left\lfloor\rho_{k}^{(\x)}\right\rfloor \log\joint{k},
    \label{eq:16}
\end{align}
where $\lfloor\cdot\rfloor$ is the \emph{stop-gradient} operator preventing gradient backpropagation, as the responsibilities $\rho^{(\x)}$ from the E-step are held constant during the M-step as in canonical EM\footnote{Stochastic gradient EM admits a convenient factorized form that will later be useful for deriving natural gradient EM in \cref{sec:3.3}.}.

\subsection{Connecting EM and NGD}
\label{sec:3.2}

Intriguingly, we observe that in practice, the negative log-likelihood loss \eqref{eq:3} and the stochastic gradient EM loss \eqref{eq:16} always result in \emph{identical} models under controlled random seeds. We can formalize this insight by proving that $\nabla\Ell_{\nll}(\theta)=\nabla\Ell_{\sgem}(\theta)$ as follows.
\begin{proposition}[\citealp{salakhutdinov2003optimization,xu2024toward}]
    \label{proof:1} %
    Consider a probabilistic model $p(\x,\z|\theta)$ with observed variables
    $\x$, latent variables $\z$, and parameters $\theta$. We can show that
    \begin{equation*}
        \nabla_{\theta}\log p(\x|\theta)\big\vert_{\theta=\theta_t}
        = \nabla_{\theta}Q(\theta | \theta_{t})\big\vert_{\theta=\theta_t},
    \end{equation*}
    where $Q(\theta | \theta_{t})$ is the M-step objective \eqref{eq:10}
    \begin{equation*}
        Q(\theta | \theta_{t}) =
        \int p(\z|\x;\theta_{t}) \log p(\x,\z|\theta ) \dif\z.
    \end{equation*}
    \begin{proof}
        See \cref{app:b1}.
    \end{proof}
\end{proposition}

\cref{proof:1} states that the gradient of the observed-data log-likelihood $\log p(\x|\theta )$ is equal to the gradient of the M-step objective $Q(\theta | \theta_{t})$ at $\theta =\theta_{t}$. We can extend this conclusion to mixture density networks as follows.

\begin{corollary}
    \label{proof:2} %
    Consider a mixture density network $f(\x;\theta)$ parametrized by $\theta$. We can show that
    \begin{equation*}
        \nabla_{\theta}\Ell_{\nll}(\theta)\big\vert_{\theta=\theta_t}%
        = \nabla_{\theta}\Ell_{\sgem}(\theta)\big\vert_{\theta=\theta_t},
    \end{equation*}
    where $\theta_{t}$ is the parameter used for evaluating $\rho^{(\x)}$ \eqref{eq:15}.
    \begin{proof}
        See \cref{app:b2}.
    \end{proof}
\end{corollary}

\cref{proof:1,proof:2} show that \sgem offers no additional benefits over \nll when only first-order optimizers (e.g. stochastic GD, Adam \cite{kingma2017adam}) are involved, as their gradients are equal.

However, \sgem unlocks an alternative view of \nll that allows bypassing the intractability of Gaussian mixture FIM and enables tractable natural gradient updates for MDNs. We start by showing that exact EM is equivalent to NGD.

\begin{proposition}[\citealp{sato2001online}]
    \label{proof:3} %
     For EM algorithms with exact E-M steps \textup{(\ref{eq:9},~\ref{eq:10})}, we show that the update $\theta_{t} \to \theta_{t+1}$ is equivalent to a natural gradient descent step on the negative log-likelihood up to second-order Taylor approximation
    \begin{equation*}
        \theta_{t+1} = \theta_{t} - \hat{F}(\theta_{t})^{-1} \nabla_{\theta} (-\log p(\x|\theta_{t})),
    \end{equation*}
    where the learning rate is one, and $\hat{F}(\theta_t)$ is the complete-data Fisher information matrix at $\theta=\theta_{t}$
    \begin{equation*}
        \hat{F}(\theta) = -\EE_{p(\x,\z|\theta)}[\nabla^{2}_{\theta}\log p(\x,\z|\theta)].
    \end{equation*}
    \begin{proof}
        See \cref{app:b3}
    \end{proof}
\end{proposition}

For mixture density networks $f(\x;\theta)$, although exact EM updates are not possible, we can emulate such behaviors via natural gradient descent on the \sgem objective \eqref{eq:16} using the \emph{complete-data FIM} $\hat{F}(\phi^{(\x)})$ w.r.t. the Gaussian mixture parameters $\phi^{(\x)}$ \eqref{eq:1}:
\begin{align}
    \theta_{t+1}
    & = \theta_{t} - \hat{F}(\phi^{(\x)})^{-1} \underbrace{\nabla_{\phi^{(\x)}} \Ell_{\nll}(\theta_{t}) \cdot \nabla_{\theta} \phi^{(\x)}}_{\text{Chain rule}}, \notag \\
    & = \theta_{t} - \hat{F}(\phi^{(\x)})^{-1} \underbrace{\nabla_{\phi^{(\x)}} \Ell_{\sgem}(\theta_{t}) \cdot \nabla_{\theta} \phi^{(\x)}}_{\text{\cref{proof:2}}}. %
    \label{eq:17}
\end{align}
Furthermore, we can show that the complete-data FIM of Gaussian mixtures is block-diagonal, allowing independent natural gradient updates for each component distribution.

\begin{proposition}
    \label{proof:4} %
    Consider a Gaussian mixture $p(\x,\z|\theta)$ with observed variables $\x$, latent variables $\z$, and parameters $\theta=\{\pi_{k},\mu_{k},\Sigma_{k}\}_{k=1}^{K}$. We can show that the complete-data FIM $\hat{F}(\theta)$ of Gaussian mixtures is block-diagonal with $K+1$ blocks as shown below:
    \begin{equation*}
        \hat{F}(\theta) = \begin{bmatrix}
            \pi_{1}F_{1} & \cdots & 0 & 0 \\
            \vdots & \ddots & \vdots & \vdots \\
            0 & \cdots & \pi_{K}F_{K} & \vdots \\
            0 & \cdots & \cdots & F_{\pi}
        \end{bmatrix},
    \end{equation*}
    where $F_{k}$ for $k\in\{1,\dots,K\}$ is the FIM of the $k$-th Gaussian, and $F_{\pi}$ is the FIM of the categorical distribution.
    \begin{proof}
        See \cref{app:b4}
    \end{proof}
\end{proposition}

\cref{proof:4,eq:17} enable tractable natural gradient EM updates for MDNs, which we detail in \cref{sec:3.3}.

\begin{tcolorbox}[
    arc=3pt,
    colback=gray!5,
    colframe=white!50!black!50,
    rounded corners,
]
\textbf{Summary} \space
We prove that \emph{(a)} \sgem is equivalent to \nll in gradients, and \emph{(b)} exact EM is \mbox{equivalent} to NGD. We proceed to show that one can construct tractable natural gradient EM updates by combining \sgem and the block-diagonal complete-data FIM.
\begin{center}
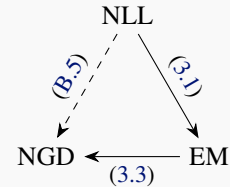

\begin{tikzpicture}[
    scale=0.5,
    arrow/.style={-{Stealth[scale=1.2]}}
]
    \node (NLL) at (90:2.5) {NLL};
    \node (NGD) at (210:2.5) {NGD};
    \node (EM)  at (330:2.5) {EM};
    \draw[arrow, dashed] (NLL) -- (NGD)
        node[midway, sloped, above, font=\footnotesize] {$\eqref{app:b5}$};
    \draw[arrow] (NLL) -- (EM)
        node[midway, sloped, above, font=\footnotesize] {$\eqref{proof:1}$};
    \draw[arrow] (EM)  -- (NGD)
        node[midway, below, font=\footnotesize] {$\eqref{proof:3}$};
\end{tikzpicture}
\captionsetup{hypcap=false}
\captionof{figure}{Key concepts and connections in \cref{sec:3.2}.}
\end{center}
\end{tcolorbox}

\subsection{Natural Gradient EM}
\label{sec:3.3}

Drawing upon previous theoretical analyses (\cref{sec:3.2}), we show in this section how one can combine the \sgem objective and the complete-data FIM $\hat{F}(\phi^{(\x)})$ of Gaussian mixtures \eqref{eq:17} to derive tractable natural gradient EM updates for MDNs. Specifically, we leverage the block-diagonal structure of $\hat{F}(\phi^{(\x)})$, which essentially factorizes an NGD update on Gaussian mixture parameters $\phi^{(\x)}$ into $K+1$ independent NGD updates on each categorical or Gaussian distribution.

To motivate the factorization perspective, we first rewrite the \sgem objective \eqref{eq:16} into the following identical but factorized \ngem objective:
\begin{align}
    \Ell_{\ngem}(\theta) = & H\left(\sg{\rho^{(\x)}}, \pi^{(\x)}\right) \notag  \\
                         - & \sum_{k=1}^{K}\sg{\rho_{k}^{(\x)}}\log\mathcal{N}(\y;\mu_{k}^{(\x)},\Sigma_{k}^{(\x)}), %
    \label{eq:18}
\end{align}
where the first term is the cross-entropy between $\rho^{(\x)}$ and $\pi^{(\x)}$, and the other $K$ terms are the negative log-likelihoods of the Gaussian components weighted by $\textstyle \rho_{k}^{(\x)}$.

The \ngem objective builds on \sgem by preconditioning the gradients with the complete-data FIM. We denote the gradient of the \ngem objective w.r.t. $\phi^{(\x)}$ as
\begin{equation}
    \nabla_{\phi^{({\x})}} =
    \begin{bmatrix}
        \nabla_{1} & \cdots & \nabla_{K} & \nabla_{\pi}
    \end{bmatrix}^{\top}, %
    \label{eq:19}
\end{equation}
where $\nabla_{k}$ for $k\in\{1,\dots,K\}$ is the gradient of the $k$-th Gaussian, and $\nabla_{\pi}$ is the gradient of categorical distribution. Therefore, the complete-data natural gradient w.r.t. $\phi^{(\x)}$ is
\begin{align}
    \hat{\nabla}_{\phi^{(\x)}} & = \hat{F}(\phi^{(\x)})^{-1} \nabla_{\phi^{(\x)}} \notag \\
    & = \begin{bmatrix}
        \dfrac{F_{1}^{-1}}{\pi_{1}}\nabla_{1} & \cdots & \dfrac{F_{K}^{-1}}{\pi_{K}}\nabla_{K} & F_{\pi}^{-1}\nabla_{\pi}
    \end{bmatrix}^{\top},
    \label{eq:20}
\end{align}
due to the block-diagonal structure of $\hat{F}(\phi^{(\x)})$.

We thus show that the complete-data natural gradient $\hat{\nabla}_{\phi^{(\x)}}$, which is core to the natural gradient EM updates \eqref{eq:17}, can be easily computed by preconditioning the regular gradient $\nabla_{\phi^{({\x})}}$ with the inverse of FIMs of individual distributions. For the exponential family distributions (including Gaussian and categorical), \citet{khan2023bayesian} demonstrate that their natural gradients are often \emph{easy} to compute.

For \textbf{Gaussian} distributions, we adopt the common assumption of diagonal covariances in deep learning \cite{kingma2013auto}. We show in \cref{app:b6} that for Gaussians parametrized by the mean $\mu$ and (diagonal) standard deviation $\sigma$, we have the following analytical \emph{diagonal} FIM
\begin{equation}
    F(\begin{bmatrix}\mu & \sigma\end{bmatrix}) =
    \begin{bmatrix}
        \dfrac{1}{\sigma^{2}} & 0 \\
        0 & \dfrac{2}{\sigma^{2}}
    \end{bmatrix}. %
    \label{eq:21}
\end{equation}

For \textbf{categorical} distributions parametrized by the logits $\psi$ such that probabilities $\pi=\mathrm{softmax}(\psi)$, we also derive in \cref{app:b7} that the analytical FIM is
\begin{equation}
    F(\psi) = \mathrm{diag}(\pi) - \pi\pi^{\top}, %
    \label{eq:22}
\end{equation}
which is, however, rank-deficient ($\mathrm{rank}(F(\psi))=K-1$) and hence not directly invertible. We replace $F(\psi)^{-1}$ with the Moore-Penrose pseudo-inverse in practice.

We have so far derived analytical FIMs for categorical and Gaussian distributions (with specific parametrization). We now present the complete natural gradient EM algorithm for optimizing mixture density networks as follows\footnote{Note that we use a learning rate $\beta<1$ in practice for stability.}:
\begin{algorithm}
\caption{Natural Gradient Expectation Maximization}
\label{algorithm:1}
\textbf{Initialize}: Training set $\mathcal{D}=\{\x_{n},\y_{n}\}_{n=1}^{N}$, mixture density network $f(\x;\theta_{0})$, learning rate $\beta$, maximum iterations $T$.

\begin{algorithmic}[1]
\FOR{each iteration $t\in\{1,\dots,T\}$}
\STATE Sample $(\x,\y)\sim\mathcal{D}$ from the training set
\STATE Compute $\phi^{(\x)}=\{\pi_{k}^{(\x)}, \mu_{k}^{(\x)}, \sigma_{k}^{(\x)}\}_{k=1}^{K}=f(\x;\theta_{t})$
\STATE (\textbf{E-step}) Compute the responsibilities $\rho^{(\x)}$ \eqref{eq:15}
$$\rho_{k}^{(\x)}=\dfrac{\pi_{k}^{(\x)}\mathcal{N}(\y;\mu_{k}^{(\x)},\sigma_{k}^{(\x)})}{\sum_{j=1}^{K}\pi_{j}^{(\x)}\mathcal{N}(\y;\mu_{j}^{(\x)},\sigma_{j}^{(\x)})}$$
\item[] (\textbf{M-step})
\STATE Compute regular gradients $\nabla_{\phi^{(\x)}}$ \eqref{eq:19} of $\Ell_{\ngem}$ \eqref{eq:18}
\STATE Compute the FIMs $F_{1},\dots,F_{K},F_{\pi}$ of Gaussian \eqref{eq:21} and categorical \eqref{eq:22} distributions
\STATE Compute complete-data natural gradients $\hat{\nabla}_{\phi^{(\x)}}$ \eqref{eq:20}
\STATE Update the mixture density network parameters $\theta$ via chain rule and backpropagation
$$\theta_{t+1}\leftarrow\theta_{t}-\beta\hat{\nabla}_{\phi^{(\x)}}\nabla_{\theta}\phi^{(\x)}$$
\ENDFOR
\end{algorithmic}
\textbf{Return}: optimized mixture density network $f(\x;\theta_{T})$
\end{algorithm}

\cref{algorithm:1} can be easily extended to the mini-batch case by summing or averaging $\hat{\nabla}_{\phi^{(\x)}}$ across samples $\x$. We also provide example implementation of the algorithm in JAX \cite{google2018jax} in \cref{app:c}, which amounts to only few lines of custom auto-differentiation hooks.

\subsection{Interpretation}
\label{sec:3.4}

In this section, we provide further justifications to intuit how \ngem exploits the information geometry to improve learning mixture density networks.

Recall that NGD penalizes the KL divergence between the distributions induced by a parameter update. As an example, consider two univariate Gaussians $p$ and $q$ with the same variance $\sigma^2$. Their KL divergence between is given by
\begin{equation}
    \KL(p\Vert q) = (\mu_{1}-\mu_{2})^{2} / (2\sigma^{2}),
\end{equation}
which implies that Gaussians with lower variances are more sensitive to parameter updates\footnote{A small change $\mu_1-\mu_2$ results in a large KL divergence.}, and vice versa. Our \ngem partially works as an automatic learning rate scheduler that scales the gradients of Gaussian components proportionally to their variances $\sigma^{2}$ \eqref{eq:21}, allowing the model to quickly jump through high-variance regions while taking conservative steps in low-variance regions, leading to faster and more stable convergence.

Furthermore, \citet{chen2024local} show that \emph{all} local minima of the Gaussian mixture \nll is a combination of \emph{(a)} one component covering multiple ground-truth Gaussians and \emph{(b)} multiples components converging to one ground-truth Gaussian. For the former case, the covering Gaussian must have high variance $\sigma^{2}$ to cover multiple modes, and \ngem allows escaping such mode collapsing local minima by again amplifying the gradients proportionally to $\sigma^{2}$.

    \section{Experiments}
\label{sec:4}

We now present experiments\footnote{Hyperparameters are provided in \cref{app:d1} for reference.} designed to answer the following key research questions:
\begin{enumerate}[align=left,leftmargin=*,noitemsep,topsep=0pt]
    \item[(\cref{sec:4.1})] Does \ngem generally yield faster convergence?
    \item[(\cref{sec:4.2})] Does \ngem help mitigate or prevent mode collapse?
    \item[(\cref{sec:4.3})] How does \ngem work with high-dimensional data?
    \item[(\cref{sec:4.4})] How does \ngem perform against baselines on real-world predictive tasks? How much computational overhead does \ngem incur?
\end{enumerate}

\subsection{Measuring Convergence}
\label{sec:4.1}

We start by measuring the convergence rates of \ngem on two synthetic examples, of which we have access to the ground-truth distributions for comparison and evaluation. The first synthetic example is the \textbf{Two-Gaussians} dataset, which contains $2n$ training data points equally sampled from two distinct and well-separated Gaussians in $\mathbb{R}^{2}$, as illustrated in \cref{fig:2a,fig:2b}.

\begin{figure}[h]
    \includegraphics[width=\linewidth]{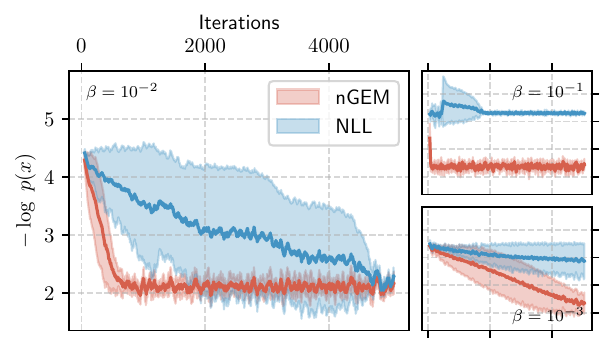}
    \caption{Negative log-likelihood ($\downarrow$) of the learned GMM on the Two-Gaussians example, with different learning rates $\beta$. Results averaged ($\pm$ std) across 5 random seeds.}
    \label{fig:4}
\end{figure}

Two-Gaussians is a simple \emph{illustrative} task that does not require a neural network to model conditional distributions. Instead we directly learn the Gaussian mixture parameters using either \ngem or \nll. \cref{fig:4} illustrates how \ngem performs against the standard \nll on this task with different learning rates $\beta$ by measuring their negative log-likelihoods. We find that \ngem consistently outperforms \nll in terms of convergence rates: \emph{(a)} \ngem can converge with only around 500 iterations while \nll takes over 5000 iterations ($\beta=10^{-2}$), and \emph{(b)} \ngem can reliably converge to the optimal solution whereas \nll catastrophically fails due to mode collapse ($\beta=10^{-1}$).

\begin{figure}[h]
    \centering
    \begin{subfigure}[h]{0.45\linewidth}
        \includegraphics[width=\linewidth]{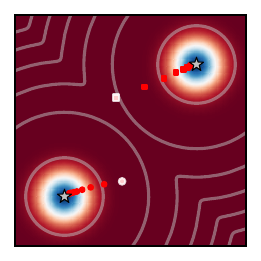}
        \caption{\ngem}
        \label{fig:5a}
    \end{subfigure}%
    \begin{subfigure}[h]{0.45\linewidth}
        \includegraphics[width=\linewidth]{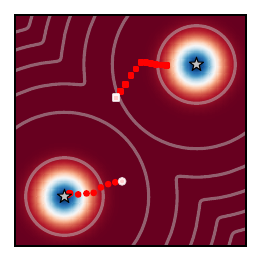}
        \caption{\nll}
        \label{fig:5b}
    \end{subfigure}
    \caption{Trajectories of the GMM ($K=2$) component means on the Two-Gaussians example during training. Stars ($\star$) represent the ground-truth Gaussian means, \textcolor{gray}{white} squares/circles represent the initial component means, and their \textcolor{red}{red} counterparts represent the trajectory of component means logged periodically during training.}
    \label{fig:5}
\end{figure}

\cref{fig:5} visualizes the trajectories of the learned Gaussian mixture component means during one training. We observe that \ngem (\cref{fig:5a}) moves along a straight path towards the targets ($\star$), presumably owing to \ngem's awareness of the information geometry. Furthermore, \ngem takes larger jumps in early iterations and switches to smaller steps near convergence, aligning with our discussion in \cref{sec:3.4}. On the other hand, \nll (\cref{fig:5b}) moves along a curved path with fixed step sizes, resulting in slower convergence.

\addtocounter{figure}{2}
\begin{figure*}[t!]
    \centering
    \begin{subfigure}[b]{0.24\textwidth}
        \includegraphics[width=\textwidth]{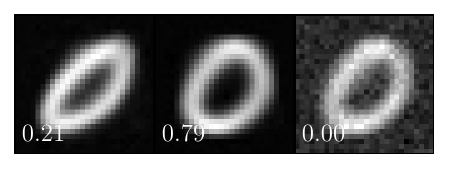}
    \end{subfigure}
    \hfill
    \begin{subfigure}[b]{0.24\textwidth}
        \includegraphics[width=\textwidth]{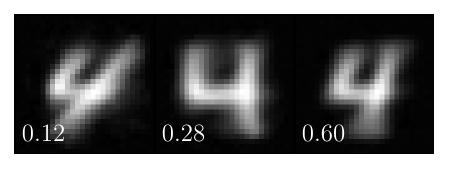}
    \end{subfigure}
    \hfill
    \begin{subfigure}[b]{0.24\textwidth}
        \includegraphics[width=\textwidth]{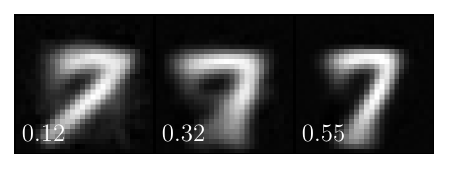}
    \end{subfigure}
    \hfill
    \begin{subfigure}[b]{0.24\textwidth}
        \includegraphics[width=\textwidth]{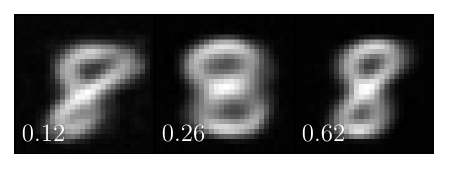}
    \end{subfigure}

    \begin{subfigure}[b]{0.24\textwidth}
        \includegraphics[width=\textwidth]{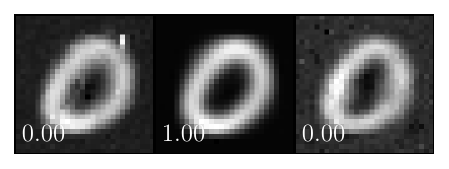}
    \end{subfigure}
    \hfill
    \begin{subfigure}[b]{0.24\textwidth}
        \includegraphics[width=\textwidth]{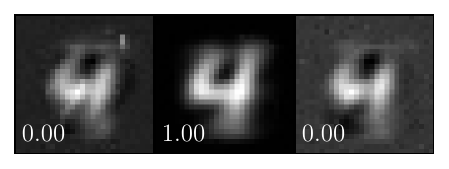}
    \end{subfigure}
    \hfill
    \begin{subfigure}[b]{0.24\textwidth}
        \includegraphics[width=\textwidth]{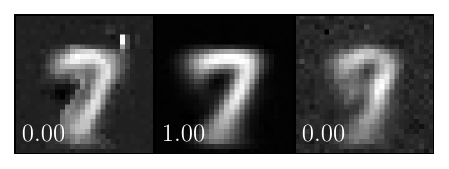}
    \end{subfigure}
    \hfill
    \begin{subfigure}[b]{0.24\textwidth}
        \includegraphics[width=\textwidth]{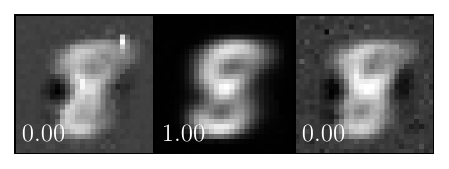}
    \end{subfigure}

    \caption{\textbf{Top row} (\ngem) v.s. \textbf{Bottom row} (\nll). Visualizing the predicted Gaussian mixture ($K=3$) component means $\mu_{k}$ given class labels $\{0,4,7,8\}$ on the inverse MNIST example. Numbers to the bottom left corner of each image denote the corresponding mixture component weight $\pi_{k}$. We refer the readers to \cref{app:d2} for full visualization results due to space constraint.}
    \label{fig:8}
\end{figure*}
\addtocounter{figure}{-3}

The second synthetic example is the \textbf{Two-Sinusoids} dataset, which requires modeling a conditional density $p(y|x)$ such that $x\sim\mathrm{Uniform}(0,4\oldpi)$ and $y$ is equally probably one of $\oldpi\sin(x)+\xi$ or $\oldpi\sin(x+\oldpi)+\xi$, with $\xi$ being an additive Gaussian noise sampled from $\mathcal{N}(0, 0.1)$. We adopt a standard multi-layer perceptron (MLP) backbone for MDN, and optimize the network parameters using either \ngem or \nll.

\begin{figure}[h]
    \includegraphics[width=\linewidth]{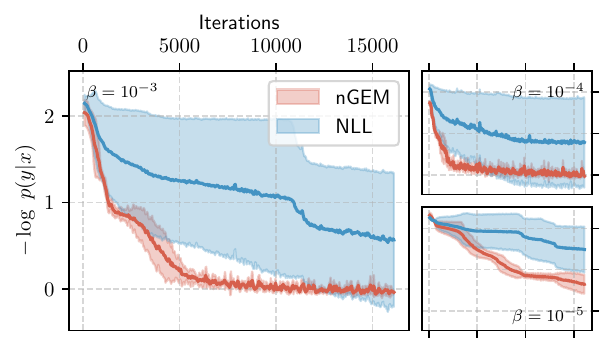}
    \caption{Negative log-likelihood ($\downarrow$) of the learned MDN on the Two-Sinusoids example, with different learning rates $\beta$. Results averaged ($\pm$ std) across 5 random seeds.}
    \label{fig:6}
\end{figure}

\cref{fig:6} illustrates how \ngem performs against standard \nll on this task with different learning rates $\beta$. We find that \ngem still generally outperforms \nll, exhibiting faster and more stable convergence across learning rates.

\begin{figure}[h]
    \centering
    \begin{subfigure}[h]{0.529\linewidth}
        \includegraphics[width=\linewidth]{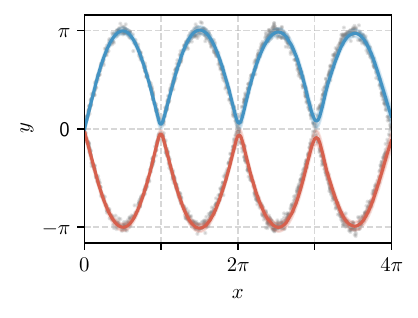}
        \caption{\ngem}
        \label{fig:7a}
    \end{subfigure}%
    \begin{subfigure}[h]{0.472\linewidth}
        \includegraphics[width=\linewidth, trim={0.2\linewidth} 0 0 0, clip]{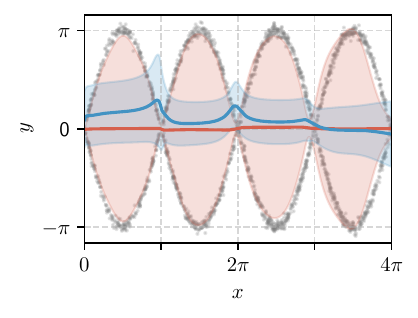}
        \caption{\nll}
        \label{fig:7b}
    \end{subfigure}
    \caption{Visualization of the learned conditional mixture density ($K=2$) on the Two-Sinusoids example for one training. \textcolor{gray}{Gray} dots in the background represent the training samples, and the \textcolor{red}{red} and \textcolor{blue}{blue} bands represent the means ($\pm$ std) of the two Gaussian components conditioned on the input $x$.}
    \label{fig:7}
\end{figure}

\cref{fig:7} demonstrates that \ngem can learn two separate conditional Gaussians to approximate the sine waves, while \nll suffers from mode collapse and only uses one high-variance Gaussian (\textcolor{red}{red}) to cover both modes.

\subsection{Measuring Mode Collapse}
\label{sec:4.2}

For evaluating mode collapse, we reuse the two synthetic examples, of which we know the optimal mixture weight is given by $\pi^{*}=[0.5, \ 0.5]$. We assess the severity of mode collapse by comparing the entropy $H(\pi)$ of the learned mixture weights against the optimal entropy $H(\pi^{*}) \approx 0.693$.

\cref{tab:1} summarizes the learned mixture weight's entropy $H(\pi)$ on Two-Gaussians with different learning rates $\beta$. Note that \ngem consistently learns near-optimal mixture weights, while \nll exhibits mode collapse for $\beta=10^{-1}$.

\begin{table}[h]
    \centering
    \caption{Entropy ($\uparrow$) of the learned mixture weight $\pi$ on the Two-Gaussians example, with different learning rates $\beta$. Results averaged ($\pm$ std) across 5 random seeds.}
    \label{tab:1}
    \begin{tabular}{@{} lccc @{}}
        \toprule
        \textbf{Method} & $\beta=10^{-1}$ & $\beta=10^{-2}$ & $\beta=10^{-3}$ \\
        \midrule
        \ngem  & \hl 0.660 $\pm$ 0.03 & \hl 0.691 $\pm$ 0.00 & \hl 0.692 $\pm$ 0.00 \\
        \nll   & 0.010 $\pm$ 0.00     & 0.687 $\pm$ 0.01     & 0.669 $\pm$ 0.02 \\
        \bottomrule
    \end{tabular}
\end{table}

\cref{tab:2} summarizes the learned mixture weight's entropy $H(\pi)$ on Two-Sinusoids with different learning rates $\beta$. For this task, \ngem can again consistently learn near-optimal mixture weights, whereas \nll is unstable (high std) and suffers from mode collapse across learning rates.

\begin{table}[h]
    \centering
    \caption{Entropy ($\uparrow$) of the learned mixture weight $\pi$ on the Two-Sinusoids example, with different learning rates $\beta$. Results averaged ($\pm$ std) across 5 random seeds.}
    \label{tab:2}
    \begin{tabular}{@{} lccc @{}}
        \toprule
        \textbf{Method} & $\beta=10^{-3}$ & $\beta=10^{-4}$ & $\beta=10^{-5}$ \\
        \midrule
        \ngem  & \hl 0.693 $\pm$ 0.00 & \hl 0.693 $\pm$ 0.00 & \hl 0.693 $\pm$ 0.00 \\
        \nll   & 0.416 $\pm$ 0.38     & 0.555 $\pm$ 0.31     & 0.413 $\pm$ 0.38 \\
        \bottomrule
    \end{tabular}
\end{table}

\subsection{Modeling High-Dimensional Data}
\label{sec:4.3}

We proceed to investigate how well \ngem scales to high-dimensional data. We consider the \textbf{inverse MNIST} problem, where one aims to predict the handwritten digit image $\x\in\mathbb{R}^{28 \times 28}$ given its class label $\y\in\{0,\dots,9\}$.

\newcommand{\m}{\phantom{-}}
\addtocounter{table}{2}
\begin{table*}[ht]
    \centering
    \caption{Negative log-likelihoods ($\downarrow$) and rooted mean squared errors (RMSE, $\downarrow$) of 10 baselines on test datasets for 3 regression tasks. Results are averaged ($\pm$ std) across 5 random seeds. Note that \emph{(a)} negative log-likelihoods are not available for MSE, \emph{(b)} RMSE for Gaussians ($\beta$-\nll) is computed between the mean and the target, i.e. $\mathrm{RMSE}(\mu,\y)$, and \emph{(c)} RMSE for Gaussian mixtures (\ngem/\nll) is computed as the minimum RMSE between each Gaussian component mean and the target, i.e. $\min_{k}\mathrm{RMSE}(\mu_{k},\y)$.}
    \label{tab:5}
    \footnotesize
    \begin{tabular}{@{} lccccccc @{}}
        \toprule
        & \multicolumn{3}{c}{Negative Log-Likelihood ($\downarrow$)} &
        & \multicolumn{3}{c}{RMSE ($\downarrow$) $\times 10^{1}$} \\
        \cmidrule{2-4} \cmidrule{6-8}
        \textbf{Method} &
            \verb|kin8nm| & \verb|energy| & \verb|housing| & &
            \verb|kin8nm| & \verb|energy| & \verb|housing| \\
        \midrule
        \parbox{2em}{\scriptsize Adam} + \ngem &
            -1.73 $\pm$ 0.04    & -3.29 $\pm$ 0.13    & -0.39 $\pm$ 0.06    & &
            0.17 $\pm$ 0.01     & 0.49 $\pm$ 0.04     & 0.81 $\pm$ 0.07 \\
        \hspace{2em} + \nll  &
            -1.67 $\pm$ 0.06    & -3.09 $\pm$ 0.24    & -0.67 $\pm$ 0.12    & &
            0.33 $\pm$ 0.03     & 0.95 $\pm$ 0.12     & 1.02 $\pm$ 0.05 \\
        \parbox{2em}{\scriptsize KFAC} + \ngem &
            -0.69 $\pm$ 0.02    & -0.43 $\pm$ 0.01    & \m1.15 $\pm$ 0.06   & &
            0.65 $\pm$ 0.03     & 1.51 $\pm$ 0.03     & 4.43 $\pm$ 0.60 \\
        \hspace{2em} + \nll  &
            -1.64 $\pm$ 0.08    & -0.96 $\pm$ 0.08    & \m1.24 $\pm$ 0.05   & &
            0.31 $\pm$ 0.02     & 2.11 $\pm$ 0.35     & 5.11 $\pm$ 0.80 \\
        \parbox{2em}{\scriptsize Muon} + \ngem &
            -1.84 $\pm$ 0.05    & -3.19 $\pm$ 0.08    & -0.19 $\pm$ 0.02    & &
            0.15 $\pm$ 0.00     & 0.53 $\pm$ 0.01     & 1.03 $\pm$ 0.11 \\
        \hspace{2em} + \nll  &
            -1.80 $\pm$ 0.06    & -2.68 $\pm$ 0.17    & -0.24 $\pm$ 0.01    & &
            0.25 $\pm$ 0.00     & 1.32 $\pm$ 0.18     & 1.51 $\pm$ 0.11 \\
        \parbox{2em}{\scriptsize Soap} + \ngem &
            \hl-1.96 $\pm$ 0.05 & \hl-4.18 $\pm$ 0.10 & -0.48 $\pm$ 0.14    & &
            \hl0.11 $\pm$ 0.01  & \hl0.30 $\pm$ 0.01  & \hl0.77 $\pm$ 0.09 \\
        \hspace{2em} + \nll  &
            -1.93 $\pm$ 0.06    & -3.99 $\pm$ 0.18    & \hl-0.79 $\pm$ 0.11 & &
            0.21 $\pm$ 0.01     & 0.72 $\pm$ 0.05     & 1.00 $\pm$ 0.09 \\
        \parbox{2em}{\scriptsize Adam} + $\beta$-\nll &
            -1.60 $\pm$ 0.02    & -2.55 $\pm$ 0.16    & -0.22 $\pm$ 0.05    & &
            0.40 $\pm$ 0.01     & 0.97 $\pm$ 0.04     & 1.65 $\pm$ 0.10 \\
        \hspace{2em} + MSE &
            ---                 & ---                 & ---                 & &
            0.39 $\pm$ 0.01     & 0.95 $\pm$ 0.02     & 1.42 $\pm$ 0.04 \\
        \bottomrule
    \end{tabular}
\end{table*}
\addtocounter{table}{-3}

\cref{fig:8} inspects the predicted Gaussian mixture ($K=3$) component means given labels. We observe that \ngem (\textbf{top row}) can use different components to capture subtle style variations and predict sharp images. On the other hand, \nll (\textbf{bottom row}) suffers from mode collapse (some $\pi_{k}=1$) and can only predict blurry, visually similar images.

\begin{table}[h]
    \centering
    \caption{Negative log-likelihood ($\downarrow$) and entropy ($\uparrow$) of the learned MDN ($K=3$) for inverse MNIST. We use the same hyperparameters and average metrics ($\pm$ std) across 5 random seeds.}
    \label{tab:3}
    \begin{tabular}{@{} lcc @{}}
        \toprule
        \textbf{Method} & $-\log p(\x|\y)$ ($\downarrow$) & $H(\pi)$ ($\uparrow$) \\
        \midrule
        \ngem  & \hl -1529.7 $\pm$ 101 & \hl 0.454 $\pm$ 0.15 \\
        \nll   & -1392.8 $\pm$ 178     & 0.207 $\pm$ 0.22    \\
        \bottomrule
    \end{tabular}
\end{table}

\cref{tab:3} summarizes the learned mixture density network's negative log-likelihood and entropy on the inverse MNIST problem. We find that overall \ngem outperforms \nll with lower negative log-likelihood and higher entropy. 

We also examine the over-parametrized setting, where we set the mixture density networks to have $K=10$ components. \cref{tab:4} shows that \ngem can perform reliably even in such challenging settings.

\begin{table}[h]
    \centering
    \caption{Negative log-likelihood ($\downarrow$) and entropy ($\uparrow$) of the learned MDN ($K=10$) for inverse MNIST. We use the same hyperparameters and average metrics ($\pm$ std) across 5 random seeds.}
    \label{tab:4}
    \begin{tabular}{@{} lcc @{}}
        \toprule
        \textbf{Method} & $-\log p(\x|\y)$ ($\downarrow$) & $H(\pi)$ ($\uparrow$) \\
        \midrule
        \ngem  & \hl -1601.4 $\pm$ 146 & \hl 0.688 $\pm$ 0.00 \\
        \nll   & -1470.0 $\pm$ 148     & 0.456 $\pm$ 0.32    \\
        \bottomrule
    \end{tabular}
\end{table}

\subsection{Ablation Studies}
\label{sec:4.4}

We conduct additional baseline experiments on three standard real-world predictive modeling tasks from the UCI datasets \cite{hernandez2015probabilistic}, involving robot kinematics (\verb|kin8nm|), energy efficiency (\verb|energy|), and Boston housing price regression (\verb|housing|).

For baselines, we consider mixture density networks optimized with \ngem/\nll and different optimizers, including Adam \cite{kingma2017adam} and other curvature-aware preconditioning gradient optimizers: KFAC \cite{martens2015kfac}, Muon \cite{liu2025muon} and Soap \cite{vyas2025soap}. We also consider $\beta$-\nll \cite{seitzer2022on}, which uses conditional \emph{Gaussian} distributions for predictive modeling, and the canonical mean squared error (MSE) for regression.

\cref{tab:5} reports the the performance of each baseline on all datasets. We summarize our findings as follows:
\begin{itemize}[topsep=0pt,noitemsep]
    \item Mixture density networks with \ngem/\nll in general outperform $\beta$-\nll and MSE.
    \item When using the same optimizer, \ngem outperforms \nll in RMSE on all tasks, and also in log-likelihoods except \verb|housing| (which is due to \ngem overfitting on the train set).
    \item The Soap optimizer combined with \ngem in general performs best, in terms of log-likelihoods and RMSE.
\end{itemize}
In particular, \cref{tab:5} highlight that \ngem is not competing with, but complementary to other preconditioning gradient optimizers. Our proposed \ngem operates in the Gaussian mixture distribution space, orthogonal to the neural network parameter space where other optimizers operate in. In fact, we find that integrating both methods yields additive performance gains that surpass either one in isolation.

\cref{tab:6} reports the wall-clock time (in seconds) of \ngem and \nll. We observe no discernible difference in runtime up to hundreds of updates, as \ngem does involve the usual expensive $O(n^{3})$ matrix inversion operations (\ref{eq:21},~\ref{eq:22}).

\addtocounter{table}{1}
\begin{table}[H]
    \centering
    \caption{Wall-clock time ($\downarrow$) of \ngem and \nll running for $10^{4}$ gradient updates with Adam optimizer on 3 regression tasks. Results are averaged ($\pm$ std) across 5 runs.}
    \label{tab:6}
    \begin{tabular}{@{}lccc@{}}
        \toprule
        \textbf{Method} & \verb|kin8nm| & \verb|energy| & \verb|housing| \\
        \midrule
        \ngem & 9.38 $\pm$ 0.08s & 9.40 $\pm$ 0.05s & 9.37 $\pm$ 0.05s \\
        \nll  & 9.33 $\pm$ 0.06s & 9.31 $\pm$ 0.08s & 9.31 $\pm$ 0.11s \\
        \bottomrule
    \end{tabular}
\end{table}

    \section{Related Work}

$\boldsymbol{\beta}$\textbf{-\textsc{NLL}} \cite{seitzer2022on} is a method for predictive uncertainty estimation by weighting the Gaussian \nll with a $\beta$-exponentiated variance $\sigma^{2\beta}$. Our proposed method \ngem decomposes into $\beta$-\nll when $K=1$ and $\beta=1$, in which case preconditioning gradients with the inverse Gaussian FIM \eqref{eq:21} is equivalent (up to constants) to weighting the Gaussian 
\nll objective with $\sigma^{2}$.

\textbf{Applications} \space Mixture density networks have grown popular for many downstream deep learning applications, for instance speech synthesis \cite{capes2017siri}, model-based reinforcement learning \cite{ha2018world}, sketch generation \cite{ha2018neural}, drone trajectory prediction \cite{makansi2019overcoming}, and human pose estimation \cite{li2019generating}. However, systematic investigations into learning objective pathologies of MDNs has remained scarce.

\textbf{Variational Inference} \space Though orthogonal to our work, Gaussian mixtures have also been used as surrogate posteriors in variational Bayesian inference \cite{lin2019fast, morningstar2021automatic, arenz2023unified}. For variational inference, one typically seeks to approximate an intractable posterior $p(\z|\x)$ by minimizing the KL divergence between a flexible surrogate $q(\z|\x)$ and the posterior. However, a key distinction is that in variational inference, one typically does not have access to samples from the target distribution for learning (in contrast to our work).
    \section{Conclusion}

In this work, we introduce natural gradient expectation maximization (\ngem) for mixture density network \mbox{optimization}. We identify theoretical connections between EM and NGD to derive tractable natural gradient EM updates for MDN. Our proposed method \ngem can stabilize and accelerate learning mixture density networks, incurs negligible computational overheads, while remaining easy to implement in practice (see \cref{app:c}).

Our empirical evaluations across a range of tasks -- from synthetic examples to high-dimensional inverse problems -- suggest that \ngem is both effective and robust. We expect \ngem to become a \emph{drop-in} alternative to the standard \nll objective for learning deep mixture density networks.

\textbf{Limitations and Future Work} \space One of the fundamental assumption we rely on is the diagonal Gaussian covariance matrix, which is, although common in deep learning, still restrictive. An interesting future research direction would be deriving tractable natural gradient EM updates for Gaussian components with low-rank covariance approximations. Another future direction would be extending \ngem to non-Gaussian mixture models within the exponential family.



    \section*{Impact Statement}
    This paper presents work whose goal is to advance the field of Machine Learning.
    There are many potential societal consequences of our work, none which we feel
    must be specifically highlighted here.

    \bibliography{references}
    \bibliographystyle{icml2026}

    \newpage
    \appendix
    \onecolumn

    \section{Natural Gradient Descent}
\label{app:a}

In machine learning, the gradient descent algorithm for minimizing some objective function $J(\theta):\mathbb{R}^n\to\mathbb{R}$ is
\begin{equation}
    \theta_{t+1} \leftarrow \theta_{t} - \beta \nabla J(\theta_t),
\end{equation}
where $\beta$ is commonly known as the \emph{learning rate}.

One can show that the above gradient descent update can be derived from minimizing the first-order Taylor expansion of $J(\theta)$ at $\theta=\theta_{t}$ with an additional Euclidean distance penalty
\begin{equation}
    \theta_{t+1} = \operatornamewithlimits{\arg\min}_{\theta}
    \underbrace{J(\theta_{t}) + \nabla J(\theta_{t})^{\top}(\theta-\theta_{t})}_{\text{Taylor expansion}}
    + (2\beta)^{-1}\underbrace{\lVert\theta-\theta_{t}\rVert_{2}^{2}}_{\text{Penalty}}.
\end{equation}
Therefore, we say that gradient descent implicitly assumes that the model parameters $\theta$ lives in a \emph{Euclidean space} by using a Euclidean distance penalty to constrain large updates. However, for probabilistic models $p(\x|\theta)$, the parameter space is generally not Euclidean. Instead, a more natural penalty term would be the Kullback-Leibler (KL) divergence between the induced distributions, resulting in the following optimization objective
\begin{equation}
    \theta_{t+1} = \operatornamewithlimits{\arg\min}_{\theta}
    J(\theta_{t}) + \nabla J(\theta_{t})^{\top}(\theta-\theta_{t})
    +\beta^{-1}\underbrace{\KL(p(\x|\theta_{t}\Vert p(\x|\theta))}_{\text{Penalty}}.
\end{equation}
Since $\KL(p\Vert q)$ does not have an analytical form generally, we instead consider its second-order Taylor expansion at $\theta=\theta_{t}$
\begin{align}
    \KL(p(\x|\theta_{t}\Vert p(\x|\theta))
    & \approx \KL(p(\x|\theta_{t}\Vert p(\x|\theta_{t})) \notag \\
    & + (\nabla_{\theta} \KL(p(\x|\theta_{t}\Vert p(\x|\theta))\big\vert_{\theta=\theta_{t}})^{\top}(\theta-\theta_{t}) \notag \\
    & + \dfrac{1}{2}(\theta-\theta_{t})^{\top} (\nabla^{2}_{\theta} \KL(p(\x|\theta_{t}\Vert p(\x|\theta))\big\vert_{\theta=\theta_{t}}) (\theta-\theta_{t}),
\end{align}
where
\begin{enumerate}
    \item The first term $\KL(p(\x|\theta_{t}\Vert p(\x|\theta_{t}))$ is trivially \emph{zero}.
    \item The second term is also \emph{zero} because the expectation of the score function $\nabla\log p(\x)$ is zero.
    \begin{equation}
        \nabla_{\theta} \KL(p(\x|\theta_{t}\Vert p(\x|\theta))\big\vert_{\theta=\theta_{t}}
        = -\EE_{p(\x|\theta_{t})}\left[\underbrace{\nabla_{\theta}\log p(\x|\theta_{t})}_{\text{Score}}\right]
        = 0.
    \end{equation}
    \item The third term is $\dfrac{1}{2}(\theta-\theta_{t})^{\top} F(\theta_{t}) (\theta-\theta_{t})$, where $F(\theta_{t})$ is the Fisher information matrix \eqref{eq:6} at $\theta=\theta_{t}$.
    \begin{equation}
        \nabla_{\theta}^{2} \KL(p(\x|\theta_{t}\Vert p(\x|\theta))\big\vert_{\theta=\theta_{t}}
        = -\EE_{p(\x|\theta_{t})}\left[\nabla_{\theta}^{2} \log p(\x|\theta_{t})\right]
        = F(\theta_{t}).
    \end{equation}
\end{enumerate}
Therefore, we have $\KL(p(\x|\theta_{t}\Vert p(\x|\theta)) \approx \dfrac{1}{2}(\theta-\theta_{t})^{\top} F(\theta_{t}) (\theta-\theta_{t})$. Plugging back the approximation and solving the new optimization objective
\begin{equation}
    \theta_{t+1} = \operatornamewithlimits{\arg\min}_{\theta}
    J(\theta_{t}) + \nabla J(\theta_{t})^{\top}(\theta-\theta_{t})
    + (2\beta)^{-1}\underbrace{(\theta-\theta_{t})^{\top} F(\theta_{t}) (\theta-\theta_{t})}_{\text{Penalty}}.
\end{equation}
lead to the following natural gradient descent (NGD) update
\begin{equation}
    \theta_{t+1} \leftarrow \theta_{t}- \beta F^{-1}(\theta_{t})\nabla J(\theta_{t}).
\end{equation}
Natural gradient descent yields the steepest descent on a Riemannian manifold with the Fisher information metric, which subsumes the standard gradient descent on the Euclidean space as a special case when $F(\theta)=I$.

\clearpage

    \section{Proofs and Derivations}
\label{app:b}

\subsection{Proof of \cref{proof:1}}
\label{app:b1}

We assume a probabilistic model $p(\x,\z|\theta)$ with observed variables $\x$, latent variables $\z$, and parameters $\theta$.

We begin by rewriting the gradient of the observed-data log likelihood as follows:
\begin{align}
    \nabla_{\theta}\log p(\x|\theta)
    & = \dfrac{1}{p(\x|\theta)}\nabla_{\theta} p(\x|\theta) \notag \\
    & = \dfrac{1}{p(\x|\theta)}\nabla_{\theta}\int p(\x,\z|\theta)\dif\z \notag \\
    & = \dfrac{1}{p(\x|\theta)}\int \nabla_{\theta}p(\x,\z|\theta)\dif\z \notag \\
    & = \dfrac{1}{p(\x|\theta)}\int p(\x,\z|\theta)\nabla_{\theta}\log p(\x,\z|\theta)\dif\z \notag \\
    & = \int p(\z|\x,\theta) \nabla_{\theta}\log p(\x,\z|\theta)\dif\z.
\end{align}

Note that by definition $Q(\theta\mid\theta_{t}) = \int p(\z|\x,\theta_{t}) \log p(\x,\z|\theta )\dif\z$. It is then clear that
\begin{equation}
    \nabla_{\theta}\log p(\x|\theta)\big\vert_{\theta=\theta_{t}}
    = \int p(\z|\x,\theta_{t}) \nabla_{\theta}\log p(\x,\z|\theta_{t})\dif\z
    = \nabla_{\theta}Q(\theta\mid\theta_{t})\big\vert_{\theta=\theta_t}.
\end{equation}

\subsection{Proof of \cref{proof:2}}
\label{app:b2}

We view the mixture density network $f(\x;\theta)$ as a latent variable model $p(\y,\z|\x;\theta)$ with observed features $\x$ and targets $\y$, latent variables $\z$, and network parameters $\theta$.

Following \cref{proof:1}, with a trivial substitution of $p(\x,\z|\theta)$ with $p(\y,\z|\x;\theta)$, we can similarly derive that
\begin{equation}
    \nabla_{\theta}\log p(\y|\x;\theta) %
    = \int p(\z|\x,\y;\theta) \nabla_{\theta}\log p(\y,\z|\x;\theta) \dif\z.
\end{equation}

Recall that by definition we have
\begin{align}
     & \Ell_{\nll}(\theta) = -\log p(\y|\x;\theta), \\
     & \Ell_{\sgem}(\theta)=-\int p(\z|\x,\y;\theta_{t}) \log p(\y,\z|\x;\theta) \dif\z,
\end{align}
where $\theta_{t}$ is the parameter used for evaluating the responsibilities $\rho^{(\x)}$ \eqref{eq:15}.

Therefore, it is clear that
\begin{align}
    \nabla_{\theta}\Ell_{\nll}(\theta)\big\vert_{\theta=\theta_t}
    = -\int p(\z|\x,\y;\theta_{t}) \nabla_{\theta}\log p(\y,\z|\x;\theta_{t}) \dif\z
    = \nabla_{\theta}\Ell_{\sgem}(\theta)\big\vert_{\theta=\theta_t}.
\end{align}

\subsection{Proof of \cref{proof:3}}
\label{app:b3}
Assuming the E-step \eqref{eq:9} is exact, we consider the second-order Taylor expansion of the M-step objective $Q(\theta|\theta_{t})$ \eqref{eq:10}
\begin{align}
    Q(\theta|\theta_{t})
    & \approx Q(\theta_{t}|\theta_{t}) \notag \\
    & + (\nabla_{\theta}Q(\theta\mid\theta_{t})\big\vert_{\theta=\theta_t})^{\top} (\theta-\theta_{t}) \notag \\
    & + \dfrac{1}{2} (\theta-\theta_{t})^{\top} (\nabla_{\theta}^{2}Q(\theta\mid\theta_{t})\big\vert_{\theta=\theta_t}) (\theta-\theta_{t}).
\end{align}

Assuming the M-step \eqref{eq:10} is also exact, the gradient of $Q(\theta|\theta_{t})$ at the maximum $\theta_{t+1}$ must be zero
\begin{equation}
    \nabla_{\theta}Q(\theta\mid\theta_{t})\big\vert_{\theta=\theta_{t+1}} = 0.
\end{equation}

Replacing $Q(\theta\mid\theta_{t})$ with its the second-order Taylor approximation, we have
\begin{equation}
    (\nabla_{\theta}Q(\theta\mid\theta_{t})\big\vert_{\theta=\theta_t})
    + (\nabla_{\theta}^{2}Q(\theta\mid\theta_{t})\big\vert_{\theta=\theta_t}) (\theta_{t+1}-\theta_{t})
    \approx 0.
\end{equation}

Note that because
\begin{itemize}
    \item $\nabla_{\theta}\log p(\x|\theta)\big\vert_{\theta=\theta_{t}} = \nabla_{\theta}Q(\theta\mid\theta_{t})\big\vert_{\theta=\theta_t}$ as shown in \cref{proof:1}, and
    \item $\nabla_{\theta}^{2}Q(\theta\mid\theta_{t})\big\vert_{\theta=\theta_t} = \EE_{p(\z|\x;\theta_{t})} \left[ \nabla_{\theta}^{2} \log p(\x,\z|\theta_{t}) \right]$ by definition,
\end{itemize}
we can equivalently write as follows
\begin{equation}
    \nabla_{\theta}\log p(\x|\theta_{t}) +
    \EE_{p(\z|\x;\theta_{t})} \left[ \nabla_{\theta}^{2} \log p(\x,\z|\theta_{t}) \right]
    (\theta_{t+1}-\theta_{t}) = 0.
\end{equation}

The term $\EE_{p(\z|\x;\theta_{t})} \left[ \nabla_{\theta}^{2} \log p(\x,\z|\theta_{t}) \right]$ is sometimes known as the \emph{local} complete-data FIM. Simply taking the expectation on both sides w.r.t. a point mass $p(\x)$ over the observed data point $\x$ yields
\begin{equation}
    \nabla_{\theta}\log p(\x|\theta_{t}) +
    \underbrace{\EE_{p(\x,\z|\theta_{t})} \left[ \nabla_{\theta}^{2} \log p(\x,\z|\theta_{t}) \right]}_{-\hat{F}(\theta_{t})}
    (\theta_{t+1}-\theta_{t}) = 0,
\end{equation}
which can be rearranged as follows
\begin{equation}
    \theta_{t+1} = \theta_{t} - \hat{F}(\theta_{t})^{-1} (-\nabla_{\theta}\log p(\x|\theta_{t})).
\end{equation}

\subsection{Proof of \cref{proof:4}}
\label{app:b4}

Assuming that the complete-data log-likelihood $\log p(\x,\z|\theta)$ is twice differentiable, \citet{lin2019fast} shows that generally the complete-data FIM $\hat{F}(\theta)$ is block-diagonal. 

In this section, we show that as a special case, a Gaussian mixture model $p(\x,\z|\theta)$ with parameters $\theta=\{\pi_{k},\mu_{k},\Sigma_{k}\}_{k=1}^{K}$ has a block-diagonal $\hat{F}(\theta)$ with exactly $K+1$ blocks. We begin by writing down the complete-data log-likelihood
\begin{equation}
    \log p(\x,\z|\theta) = \sum_{k=1}^{K} \mathbbm{1}_{\{\z=k\}} [\log\pi_{k} + \log\mathcal{N}(\x;\mu_{k},\Sigma_{k})],
\end{equation}
where $\mathbbm{1}_{\{\z=k\}}$ is the characteristic function that returns 1 when $\z=k$ and 0 otherwise.

Note that $\log p(\x,\z|\theta)$ can be equivalently rewritten as
\begin{equation}
    \log p(\x,\z|\theta)
    = \sum_{k=1}^{K} \mathbbm{1}_{\{\z=k\}} \log\pi_{k}
    + \sum_{k=1}^{K} \mathbbm{1}_{\{\z=k\}} \log\mathcal{N}(\x;\mu_{k},\Sigma_{k})
    = \ell_{\pi} + \sum_{k=1}^{K} \ell_{k},
\end{equation}
where $\ell_{\pi} = \sum_{k=1}^{K} \mathbbm{1}_{\{\z=k\}} \log\pi_{k}$ and $\ell_k = \mathbbm{1}_{\{\z=k\}} \log\mathcal{N}(\x;\mu_{k},\Sigma_{k})$.

It is then clear that for any $1 \leq i, j \leq K$ and $i \neq j$, we have
\begin{equation}
    \nabla_{\mu_{i}} \nabla_{\mu_{j}} \log p(\x,\z|\theta) =
    \nabla_{\mu_{i}} \nabla_{\mu_{j}} \ell_{i} =
    \boldsymbol{0},
\end{equation}
because $\ell_{i}$ is a constant w.r.t. $\mu_{j}$. Similarly, we can show that
\begin{alignat}{3}
    & \nabla_{\mu_{i}} \nabla_{\mu_{j}} \log p(\x,\z|\theta) && =
    \nabla_{\mu_{i}} \nabla_{\Sigma_{j}} \log p(\x,\z|\theta) && = \notag \\
    & \nabla_{\Sigma_{i}} \nabla_{\mu_{j}} \log p(\x,\z|\theta) && =
    \nabla_{\Sigma_{i}} \nabla_{\Sigma_{j}} \log p(\x,\z|\theta) && = \notag \\
    & \nabla_{\mu_{i}} \nabla_{\pi} \log p(\x,\z|\theta) && =
    \nabla_{\Sigma_{i}} \nabla_{\pi} \log p(\x,\z|\theta) && = \notag \\
    & \nabla_{\pi} \nabla_{\mu_{j}} \log p(\x,\z|\theta) && =
    \nabla_{\pi} \nabla_{\Sigma_{j}} \log p(\x,\z|\theta) && =
    \boldsymbol{0}.
\end{alignat}

Therefore, we can conclude that the Hessian of the complete-data log-likelihood is block-diagonal as all the off-diagonal blocks are zero matrices as we show above.
\begin{equation}
    \nabla_{\theta}^{2} \log p(\x,\z|\theta) =
    \begin{bmatrix}
        \nabla^{2}_{\mu_{1},\Sigma_{1}}\ell_{1} & \cdots & 0 & 0 \\
        \vdots & \ddots & \vdots & \vdots \\
        0 & \cdots & \nabla^{2}_{\mu_{K},\Sigma_{K}}\ell_{K} & \vdots \\
        0 & \cdots & \cdots & \nabla^{2}_{\pi}\ell_{\pi}
    \end{bmatrix}.
\end{equation}

Finally, note that by definition the complete-data FIM $\hat{F}(\theta) = -\EE_{p(\x,\z|\theta)} [ \nabla_{\theta}^{2} \log p(\x,\z|\theta) ]$, where
\begin{itemize}
    \item The expectation of $\nabla^{2}_{\mu_{k},\Sigma_{k}}\ell_{k}$ for $k\in\{1,\dots,K\}$ is
        \begin{equation}
            \EE_{p(\x,\z|\theta)} [\nabla^{2}_{\mu_{1},\Sigma_{1}}\ell_{1}]
            = \EE_{p(\z|\theta)} [\mathbbm{1}_{\{\z=k\}}] \cdot \EE_{p(\x|\z;\theta)} [\nabla^{2}\log\mathcal{N}(\x;\mu_{k},\Sigma_{k})]
            = \pi_{k}\cdot-F_{k},
        \end{equation}
    \item The expectation of $\nabla^{2}_{\pi}\ell_{\pi}$ is
        \begin{equation}
            \EE_{p(\x,\z|\theta)} [\nabla^{2}_{\pi}\ell_{\pi}]
            = \EE_{p(\x|\z;\theta)} \EE_{p(\z|\theta)} [\nabla^{2}\log\mathrm{Cat}(\z;\pi)]
            = -F_{\pi},
        \end{equation}
\end{itemize}
with $F_{k}$ for $k\in\{1,\dots,K\}$ being the FIM of the $k$-th Gaussian and $F_{\pi}$ being the FIM of the categorical distribution.

Therefore, the complete-data FIM $\hat{F}(\theta)$ of a Gaussian mixture is given as follows
\begin{equation}
    \hat{F}(\theta) =
    -\EE_{p(\x,\z|\theta)} [ \nabla_{\theta}^{2} \log p(\x,\z|\theta) ] =
    \begin{bmatrix}
        \pi_{1}F_{1} & \cdots & 0 & 0 \\
        \vdots & \ddots & \vdots & \vdots \\
        0 & \cdots & \pi_{K}F_{K} & \vdots \\
        0 & \cdots & \cdots & F_{\pi}
    \end{bmatrix}.
\end{equation}

\subsection{Fisher Information Matrix of Gaussian Mixtures}
\label{app:b5}

Consider a Gaussian mixture model with parameters $\theta=\{\pi_{k},\mu_{k},\Sigma_{k}\}_{k=1}^{K}$. The log density is
\begin{equation}
    \log p(\x|\theta) = \log \sum_{k=1}^{K} \pi_{k} \mathcal{N}(\x;\mu_{k},\Sigma_{k}).
\end{equation}

The gradient of the log density is
\begin{equation}
    \nabla_{\theta} \log p(\x|\theta) =
    \dfrac{\sum_{k=1}^{K} \nabla_{\theta} [\pi_{k} \mathcal{N}(\x;\mu_{k},\Sigma_{k})]}{\sum_{k=1}^{K} \pi_{k} \mathcal{N}(\x;\mu_{k},\Sigma_{k})}.
\end{equation}

We can see that the gradient w.r.t. any parameter will also depend on all other parameters through the denominator term $\sum_{k=1}^{K} \pi_{k} \mathcal{N}(\x;\mu_{k},\Sigma_{k})$, which is not factorizable due to the log-sum term in $\log p(\x|\theta)$.

The interdependence structure of the log density gradient results in highly complex Hessian matrices, which is often undesirable and practically intractable to evaluate. Furthermore, one needs to compute the FIM (negative expected Hessian) via numerical integration, which exacerbates the intractability.

\subsection{Fisher Information Matrix of Gaussian Distributions}
\label{app:b6}

Consider a Gaussian distribution with parameters $\theta=\{\mu,\sigma\}$. The log density is
\begin{equation}
    \log p(\x|\theta) = -\log\det(\mathrm{diag}(\sigma))-\dfrac{1}{2}(\x-\mu)^{\top}\mathrm{diag}(\sigma^{-2})(\x-\mu).
\end{equation}

The gradient of the log density is
\begin{align}
    & \nabla_{\mu} \log p(\x|\theta) = \dfrac{\x-\mu}{\sigma^{2}}, \\
    & \nabla_{\sigma} \log p(\x|\theta) = \dfrac{(\x-\mu)^{2}}{\sigma^{3}} - \dfrac{1}{\sigma}.
\end{align}

The Hessian of the log density is
\begin{align}
    & \nabla_{\mu} \nabla_{\mu} \log p(\x|\theta) = -\dfrac{1}{\sigma^{2}}, \\
    & \nabla_{\mu} \nabla_{\sigma} \log p(\x|\theta) = -\dfrac{2(\x-\mu)}{\sigma^{3}}, \\
    & \nabla_{\sigma} \nabla_{\mu} \log p(\x|\theta) = -\dfrac{2(\x-\mu)}{\sigma^{3}}, \\
    & \nabla_{\sigma} \nabla_{\sigma} \log p(\x|\theta) = -\dfrac{3(\x-\mu)^{2}}{\sigma^{4}} + \dfrac{1}{\sigma^{2}}.
\end{align}

Note that because $\EE_{p(\x|\theta)}[\x-\mu] = 0$ and $\EE_{p(\x|\theta)}[(\x-\mu)^{2}] = \sigma^{2}$, we can derive that
\begin{equation}
    F(\begin{bmatrix}\mu & \sigma\end{bmatrix}) =
    -\EE_{p(\x|\theta)} \begin{bmatrix}
        \nabla_{\mu} \nabla_{\mu} \log p(\x|\theta) &
        \nabla_{\mu} \nabla_{\sigma} \log p(\x|\theta) \\
        \nabla_{\sigma} \nabla_{\mu} \log p(\x|\theta) &
        \nabla_{\sigma} \nabla_{\sigma} \log p(\x|\theta) 
    \end{bmatrix} =
    \begin{bmatrix}
        \dfrac{1}{\sigma^{2}} & 0 \\
        0 & \dfrac{2}{\sigma^{2}}
    \end{bmatrix}.
\end{equation}

\subsection{Fisher Information Matrix of Categorical Distributions}
\label{app:b7}

Consider a categorical distribution parametrized by logits $\psi$ such that the probabilities $\pi=\mathrm{softmax}(\psi)$. Let $\x$ be a one-hot vector describing the event outcome. The log density is
\begin{equation}
    \log p(\x|\psi) = \x^{\top}\psi - \log \sum_{k} \exp(\psi_{k}).
\end{equation}
The gradient of the log-density is
\begin{equation}
    \nabla_{\psi} \log p(\x|\theta) = \x - \pi.
\end{equation}
The Hessian of the log-density (for $i\neq j$) is
\begin{align}
    & \nabla_{\psi_{i}} \nabla_{\psi_{j}} \log p(\x|\theta) =
    - \nabla_{\psi_{j}} \pi_{i} 
    = \pi_{i}\pi_{j}, \\
    & \nabla_{\psi_{i}} \nabla_{\psi_{i}} \log p(\x|\theta) =
    - \nabla_{\psi_{i}} \pi_{i} 
    = \pi_{i}^{2} - \pi_{i}.
\end{align}

Putting together in matrix form, the Hessian
$\nabla^2 \log p(\x|\psi) = \pi\pi^{\top} - \mathrm{diag}(\pi)$. Therefore, we can derive the FIM
\begin{equation}
    F(\psi) = -\EE_{p(\x|\psi)}[\nabla^2 \log p(\x|\psi)] = \mathrm{diag}(\pi) - \pi\pi^{\top}.
\end{equation}

However, note that under this parametrization, the FIM $F(\psi)$ is not invertible (one can verify that $F(\psi)\mathbf{1}=0$). Instead, we replace the inverse of $F(\psi)$ with the Moore-Penrose pseudo-inverse in practice
\begin{equation}
    F(\psi)^{+} = \mathrm{diag}(\pi)^{-1} - \mathbf{1}\mathbf{1}^{\top}.
\end{equation}
Therefore, the complete-data natural gradient $\hat{\nabla}_{\psi}$ w.r.t. the logits $\psi$ can be simplified to
\begin{align}
    \hat{\nabla}_{\psi} 
    & = F(\psi)^{+} \nabla_{\psi} \log p(\x|\theta) \notag \\
    & = (\mathrm{diag}(\pi)^{-1} - \mathbf{1}\mathbf{1}^{\top}) (\x - \pi) \notag \\
    & = \mathrm{diag}(\pi)^{-1}\x - \mathbf{1}.
\end{align}
    \section{Implementation}
\label{app:c}
We provide a reference implementation of our approach using the JAX \cite{google2018jax} framework.

\begin{minipage}{1.0\linewidth}
\begin{verbatim}
import jax
from jax.numpy import jnp

Dist = tuple["weight", "mu", "sigma"]

def mdn(x) -> Dist: 
    ...

@jax.custom_jvp
def precond(dist: Dist) -> Dist:
    return dist

@precond.defjvp
def _(primals, tangents):
    dist, grads = *primals, *tangents
    d_w, d_mu, d_sigma = dist
    g_w, g_mu, g_sigma = grads
    return dist, (g_w, g_mu * d_std ** 2, g_std / 2.0 * d_std ** 2)

def ngem_loss(x, y):
    d_w, d_mu, d_sigma = jax.vmap(mdn)(x)
    # Apply natural gradient preconditioning
    d_w, d_mu, d_sigma = precond((d_w, d_mu, d_sigma))
    # [E-step] compute responsibilities rho
    log_weights = jax.nn.log_softmax(d_w, axis=-1)
    # Standard Gaussian log probability for each component
    log_normal = jax.vmap(gaussian_log_prob)(d_mu, d_sigma, y[:, jnp.newaxis])
    log_rho = jax.nn.log_softmax(log_weights + log_normal, axis=-1)
    rho = jax.lax.stop_gradient(jnp.exp(log_rho))
    # [M-step] natural gradient descent
    pi = jax.lax.stop_gradient(jnp.exp(log_weights))
    return -(rho / pi * (log_weights + log_normal)).sum(axis=-1).mean()
\end{verbatim}






\end{minipage}
    \section{Experiments}

We provide additional experiment details in this section. All experiments are conducted on a RTX 4090 GPU workstation.

\subsection{Hyperparameters}
\label{app:d1}

\cref{tab:7,tab:8,tab:9} reports the hyperparameter settings for all experiments discussed in \cref{sec:4}.

\begin{table}[ht]
    \centering
    \caption{Hyperparameter settings for Two-Gaussians experiments.}
    \label{tab:7}
    \setlength{\tabcolsep}{30pt}
    \begin{tabular}{@{}ll@{}}
        \toprule
        \textbf{Hyperparameter}    & \textbf{Value} \\
        \midrule
        Number of Components ($K$) & 2 \\
        Optimizer                  & \verb|SGD| \\
        Learning Rate ($\beta$)    & $\{10^{-1},10^{-2},10^{-3}\}$ \\
        Epochs                     & 50 \\
        Batch Size                 & 1 \\
        Dataset Size               & 200 (2 $\times$ 100) \\
        Random Seeds               & $\{1,2,3,4,5\}$ \\
        \bottomrule
    \end{tabular}
\end{table}

\begin{table}[ht]
    \centering
    \caption{Hyperparameter settings for Two-Sinusoids experiments.}
    \label{tab:8}
    \setlength{\tabcolsep}{30pt}
    \begin{tabular}{@{}ll@{}}
        \toprule
        \textbf{Hyperparameter}    & \textbf{Value} \\
        \midrule
        Number of Components ($K$) & 2 \\
        Hidden Layer Sizes         & $[128,128,128,128]$ \\
        Activation Function        & GELU \cite{hendrycks2016gaussian} \\
        \midrule
        Optimizer                  & \verb|Adam| \\
        Learning Rate ($\beta$)    & $\{10^{-3},10^{-4},10^{-5}\}$ \\
        Epochs                     & 1000 \\
        \midrule
        Batch Size                 & 128 \\
        Dataset Size               & 2000 (2 $\times$ 1000) \\
        Random Seeds               & $\{1,2,3,4,5\}$ \\
        \bottomrule
    \end{tabular}
\end{table}

\begin{table}[ht]
    \centering
    \caption{Hyperparameter settings for inverse-MNIST experiments.}
    \label{tab:9}
    \setlength{\tabcolsep}{30pt}
    \begin{tabular}{@{}ll@{}}
        \toprule
        \textbf{Hyperparameter}    & \textbf{Value} \\
        \midrule
        Number of Components ($K$) & $\{3,10,20\}$ \\
        Hidden Layer Sizes         & $[128,128,128,128]$ \\
        Activation Function        & GELU \cite{hendrycks2016gaussian} \\
        \midrule
        Optimizer                  & \verb|Adam| \\
        Learning Rate ($\beta$)    & $3\times10^{-4}$ \\
        Epochs                     & 100 \\
        \midrule
        Batch Size                 & 256 \\
        Dataset Size               & 60000 \\
        Random Seeds               & $\{1,2,3,4,5\}$ \\
        \bottomrule
    \end{tabular}
\end{table}

\begin{table}[H]
    \centering
    \caption{Hyperparameter settings for \{\texttt{kin8nm}, \texttt{energy}, \texttt{housing}\} experiments.}
    \label{tab:10}
    \setlength{\tabcolsep}{30pt}
    \begin{tabular}{@{}ll@{}}
        \toprule
        \textbf{Hyperparameter}    & \textbf{Value} \\
        \midrule
        Number of Components ($K$) & 3 \\
        Hidden Layer Sizes         & $[128,128]$ \\
        Activation Function        & GELU \cite{hendrycks2016gaussian} \\
        \midrule
        Optimizer                  & \{\verb|Adam|, \verb|KFAC|, \verb|Muon|, \verb|Soap|\} \\
        Learning Rate ($\beta$)    & \{\verb|kin8nm|: $10^{-3}$, \space others: $10^{-4}$\} \\
        Epochs                     & \{\verb|kin8nm|: 250, \space others: 125\} \\
        \midrule
        Batch Size                 & \{\verb|kin8nm|: 128, \space \verb|energy|: 16, \space \verb|housing|: 32\} \\
        Dataset Size               & \{\verb|kin8nm|: 8192, \space \verb|energy|: 768, \space \verb|housing|: 506\} \\
        Random Seeds               & $\{1,2,3,4,5\}$ \\
        \bottomrule
    \end{tabular}
\end{table}

\subsection{Inverse-MNIST}
\label{app:d2}

\cref{fig:9,fig:10} illustrates the Gaussian mixture ($K=3$) component means predicted by mixture density networks, trained using either \ngem (\cref{fig:9}) or \nll (\cref{fig:10}). We can observe that \ngem predicts sharp and diverse images, while \nll predicts blurry, visually similar images, and suffers from mode collapse (some $\pi_{k}=1$) for all classes.

\addtocounter{figure}{1\textbf{}}
\begin{figure}[t]
    \centering
    \begin{subfigure}[b]{0.24\textwidth}
        \includegraphics[width=\textwidth]{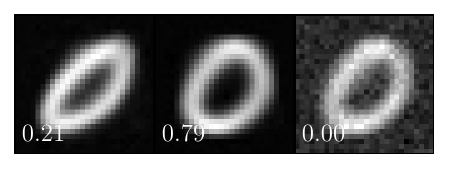}
    \end{subfigure}
    \hspace{1pt}
    \begin{subfigure}[b]{0.24\textwidth}
        \includegraphics[width=\textwidth]{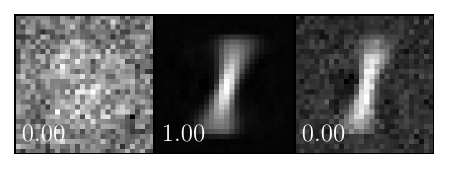}
    \end{subfigure}
    \hspace{1pt}
    \begin{subfigure}[b]{0.24\textwidth}
        \includegraphics[width=\textwidth]{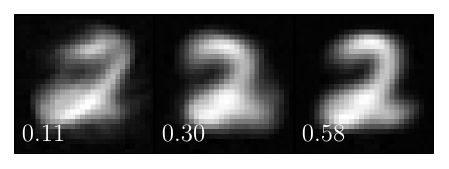}
    \end{subfigure}
    \hspace{1pt}
    \begin{subfigure}[b]{0.24\textwidth}
        \includegraphics[width=\textwidth]{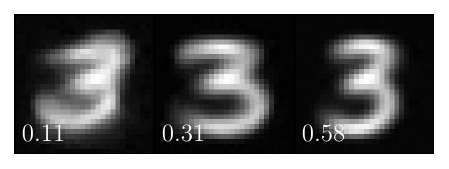}
    \end{subfigure}

    \begin{subfigure}[b]{0.24\textwidth}
        \includegraphics[width=\textwidth]{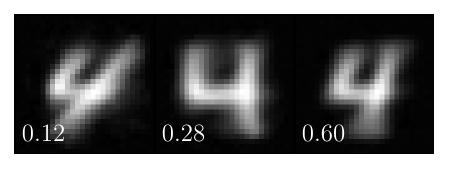}
    \end{subfigure}
    \hspace{1pt}
    \begin{subfigure}[b]{0.24\textwidth}
        \includegraphics[width=\textwidth]{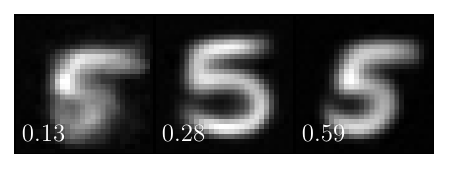}
    \end{subfigure}
    \hspace{1pt}
    \begin{subfigure}[b]{0.24\textwidth}
        \includegraphics[width=\textwidth]{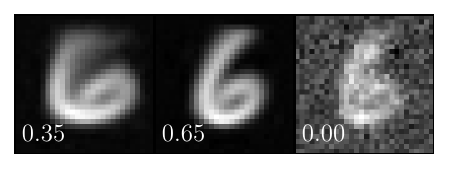}
    \end{subfigure}
    \hspace{1pt}
    \begin{subfigure}[b]{0.24\textwidth}
        \includegraphics[width=\textwidth]{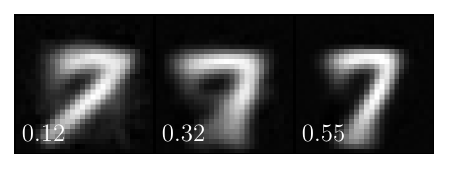}
    \end{subfigure}

    \begin{subfigure}[b]{0.24\textwidth}
        \includegraphics[width=\textwidth]{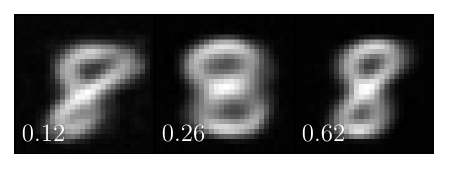}
    \end{subfigure}
    \hspace{1pt}
    \begin{subfigure}[b]{0.24\textwidth}
        \includegraphics[width=\textwidth]{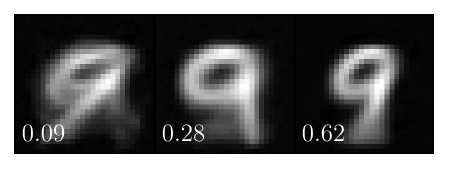}
    \end{subfigure}

    \caption{(\ngem) Visualizing the predicted Gaussian mixture ($K=3$) component means $\mu_{k}$ given class labels $\{0,1,\dots,9\}$ for the inverse-MNIST example. Numbers to the bottom left corner of each image denote the corresponding mixture component weight $\pi_{k}$.}
    \label{fig:9}
\end{figure}

\begin{figure}[t]
    \centering
    \begin{subfigure}[b]{0.24\textwidth}
        \includegraphics[width=\textwidth]{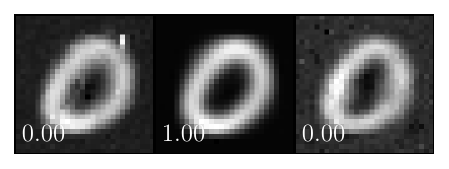}
    \end{subfigure}
    \hspace{1pt}
    \begin{subfigure}[b]{0.24\textwidth}
        \includegraphics[width=\textwidth]{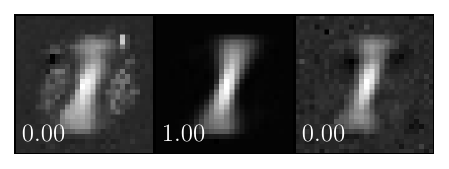}
    \end{subfigure}
    \hspace{1pt}
    \begin{subfigure}[b]{0.24\textwidth}
        \includegraphics[width=\textwidth]{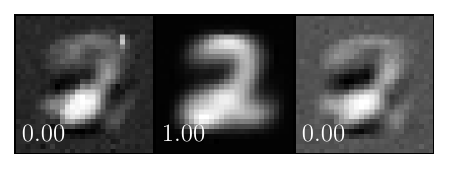}
    \end{subfigure}
    \hspace{1pt}
    \begin{subfigure}[b]{0.24\textwidth}
        \includegraphics[width=\textwidth]{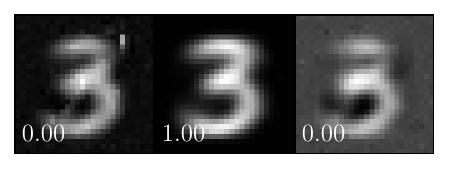}
    \end{subfigure}

    \begin{subfigure}[b]{0.24\textwidth}
        \includegraphics[width=\textwidth]{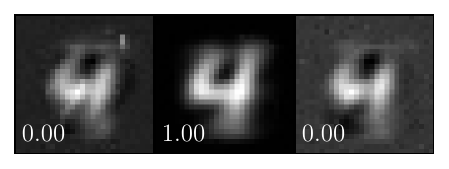}
    \end{subfigure}
    \hspace{1pt}
    \begin{subfigure}[b]{0.24\textwidth}
        \includegraphics[width=\textwidth]{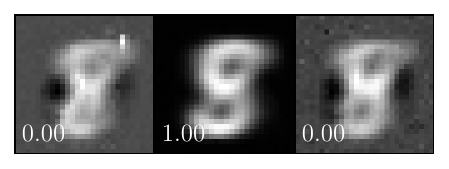}
    \end{subfigure}
    \hspace{1pt}
    \begin{subfigure}[b]{0.24\textwidth}
        \includegraphics[width=\textwidth]{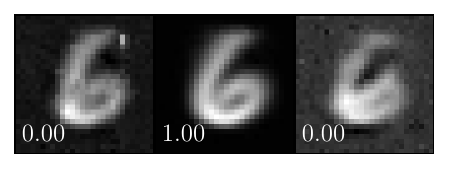}
    \end{subfigure}
    \hspace{1pt}
    \begin{subfigure}[b]{0.24\textwidth}
        \includegraphics[width=\textwidth]{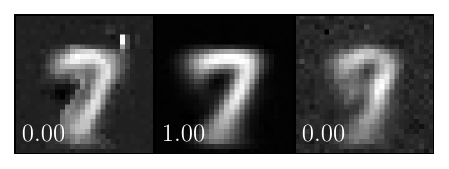}
    \end{subfigure}

    \begin{subfigure}[b]{0.24\textwidth}
        \includegraphics[width=\textwidth]{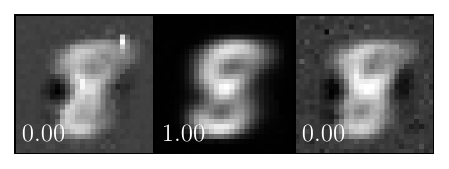}
    \end{subfigure}
    \hspace{1pt}
    \begin{subfigure}[b]{0.24\textwidth}
        \includegraphics[width=\textwidth]{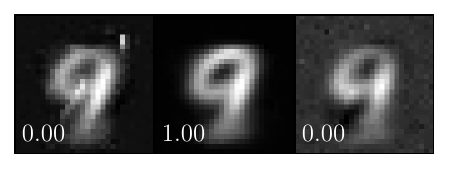}
    \end{subfigure}

    \caption{(\nll) Visualizing the predicted Gaussian mixture ($K=3$) component means $\mu_{k}$ given class labels $\{0,1,\dots,9\}$ for the inverse-MNIST example. Numbers to the bottom left corner of each image denote the corresponding mixture component weight $\pi_{k}$.}
    \label{fig:10}
\end{figure}

\raggedbottom
\end{document}